\documentclass[a4paper]{llncs}

\usepackage{moreverb}
\usepackage{algorithm}
\usepackage{algorithmic}
\usepackage{graphicx}
\usepackage[table]{xcolor}
\usepackage{multirow}
\usepackage{booktabs}
\usepackage{cite}

\usepackage{geometry}
\newcommand{\myparagraph}[1]{\noindent \textbf{#1}}
\newcommand{\etal}{\textit{et al.} }

\newcommand{\tabsize}{\fontsize{9}{9.5pt}\selectfont}

\begin{document}


\authorrunning{F.M. Castro et al.}
\titlerunning{Energy-based Tuning of Convolutional Neural Networks on Multi-GPUs}

\title{Energy-based Tuning of Convolutional Neural Networks on Multi-GPUs}

\author{Francisco M. Castro\inst{1} \and Nicol\'as Guil\inst{1} \and Manuel J. Mar\'in-Jim\'enez\inst{2} \and Jes\'us P\'erez-Serrano\inst{1} \and M. Ujald\'on\inst{1}}

\institute{Computer Architecture Department, University of Malaga, Spain \and
Department of Computing and Numerical Analysis, University of Cordoba, Spain}

\maketitle

\begin{abstract}
Deep Learning (DL) applications are gaining momentum in the realm of Artificial Intelligence, 
particularly after GPUs have demonstrated remarkable skills for accelerating their challenging 
computational requirements. Within this context, Convolutional Neural Network (CNN) models 
constitute a representative example of success on a wide set of complex applications, 
particularly on datasets where the target can be represented through a hierarchy of local features 
of increasing semantic complexity. In most of the real scenarios, the roadmap to improve results 
relies on CNN settings involving brute force computation, and researchers have lately proven 
Nvidia GPUs to be one of the best hardware counterparts for acceleration. Our work complements 
those findings with an energy study on critical parameters for the deployment of CNNs on flagship 
image and video applications: object recognition and people identification by gait, respectively. 
We evaluate energy consumption on four different networks based on the two most popular ones 
(ResNet/AlexNet): ResNet (167 layers), a 2D CNN (15 layers), a CaffeNet (25 layers) and a 
ResNetIm (94 layers) using batch sizes of 64, 128 and 256, and then correlate those with speed-up 
and accuracy to determine optimal settings. 
Experimental results on a multi-GPU server endowed with twin Maxwell and twin Pascal Titan X GPUs 
demonstrate that energy correlates with performance and that Pascal may have up to 40\% gains 
versus Maxwell. Larger batch sizes extend performance gains and energy savings, but we have 
to keep an eye on accuracy, which sometimes shows a preference for small batches. 
We expect this work to provide a preliminary guidance for a wide set of CNN and DL applications 
in modern HPC times, where the GFLOPS/w ratio constitutes the primary goal.
\end{abstract}



\section{Introduction}\label{s:intro}

We are witnessing a revolution in computer vision with the advent of Deep Learning (DL) 
architectures~\cite{bengio2015book}. Computer vision problems have been traditionally 
solved using hand-crafted features specifically designed to tackle particular 
problems~\cite{lowe1999sift, wang2011cvpr, castro2016ijprai, castro2016mva}, 
where the main challenge was to find the right descriptors for certain image contents. 
DL introduced a general way to proceed via supervised learning. 
Fukushima \etal~\cite{fukushima1980neocognitron} were pioneers developing a hierarchical 
architecture for handwritten character recognition and other pattern recognition, 
which we may consider the inspiration for Convolutional Neural Networks (CNNs). 

In 1998, LeCun \etal \cite{lecun1998lenet} introduced one of the first and most popular 
architectures for handwritten character recognition, and a decade later, 
Serre \etal~\cite{serre2008pami} contributed with a new general framework for 
the recognition of complex visual scenes. Those first steps were based on a small number 
of layers and limited datasets due to the modest computational power available, 
so researchers often moved to less demanding approaches like SVM~\cite{cortes1995svm}.

In 2012, Krizhevsky \etal \cite{krizhevsky2012nips} released `AlexNet', a CNN composed 
of 25 layers and around 60 million parameters. GPUs were capable to train the model 
with CUDA in a reasonable amount of time using four GPUs, and since then, the fascinating 
evolution of GPU performance and its recent emphasis on DL has propelled those models 
to gain extraordinary popularity.
Meanwhile, new datasets~\cite{ILSVRC15, krizhevsky2009cifar, sami20168m} containing 
millions of samples were released to train models with even more parameters without 
overfitting, promoting CNN models to be established as the state-of-art in computer vision. 
The challenge for researchers to tune computer vision applications at this point 
is no longer based on low-level features, but on general neural network components 
like number of layers, set of parameters or batch size. Within this trend, 
the last couple of years have been prolific in assorted areas like 
image recognition~\cite{szegedy2015cvpr, he2015iccv}, 
action recognition~\cite{simonyan2014nips, wang2015cvpr}, 
object detection~\cite{ren2015nips, liu2016eccv}, and 
biometric identification~\cite{marin2017icip, castro2017biosig}, 
just to mention a few akin to that of this work. 

This trend has been lately fortified with the arrival of deep learning frameworks 
publicly available, like Caffe~\cite{jia2014caffe}, 
TensorFlow~\cite{tensorflow2015-whitepaper}, CNTK~\cite{seide2016CNTK}, 
MatConvNet~\cite{vedaldi2015matconvnet} and PyTorch~\cite{paszke2017automatic}. 
Most of these frameworks are optimized for GPUs and still require large execution times, 
so energy consumption on GPUs becomes critical. That way, the flagship metric 
is no longer GFLOPS (Giga Floating-Point Operations Per Second), 
but GFLOPS/w (GFLOPS per watt). This paper emphasizes energy over speed, 
choosing representative CNN instances to shed some light about the way energy 
is spent within CNN depending on its architecture (ResNet/CaffeNet/2D-CNN), 
input dataset (images/videos) and batch size (64/128/256). Finally, a correlation 
with performance and accuracy is performed to complete our analysis.

On the hardware side, latest generations of Nvidia GPUs, namely Maxwell (2014) 
and Pascal (2016), have been used for our experimental setup. Those two generations 
have contributed like no other before to optimize the GFLOPS/w ratio, and the advantage 
amplifies in supercomputers to populate the green500 list \cite{Green500list}.
Our work gathers results combining the best GeForce model for those two generations,
Titan X, and a multi-GPU server endowed with up to four GPUs, returning somehow to 
the departure point where AlexNet emerged five years ago.

Previous works have contributed with performance analysis of DL networks in 
GPU architectures~ \cite{shi2016ccbd, kim2017spass, dong2018icpe}. We extend 
those results to energy for a more complete study using a probe plugged 
to the GPU that measures power consumption at real-time for every stage 
a deep learning algorithm consists of. Major contributions of this paper 
on DL algorithms are the following:
\begin{itemize}
    \item A combined energy and performance analysis on a multi-GPU setup 
    using the two most popular types of CNNs, and particularized for the forward, 
    backward and weight update stages of a DL algorithm.
    \item Accuracy statistics to find out the best algorithm parametrization 
    depending on three different metrics: time, energy consumption and energy-delay product.  
    \item Comparison between Maxwell and Pascal architectures for all those features above.
\end{itemize}

The rest of this paper is structured as follows.
Section \ref{s:related} introduces some related work.
Section \ref{s:CNN-overview} provides a general overview of CNNs, and 
Section \ref{s:CNN-selection} particularizes our selection of CNNs for the 
experimental study.
Section \ref{s:GPU-impl} outlines our CNN implementation on multi-GPU environments.
Section \ref{s:monitoring-energy} describes the infrastructure we have used 
for measuring energy on GPUs.
Section \ref{s:datasets} introduces the input datasets. Section \ref{s:exper-results} 
presents and discusses the experimental results, and finally, Section \ref{s:conclusions} 
summarizes the conclusions drawn from this work.


\section{Related Work}\label{s:related}

Energy consumption has gained relevance among researchers during the big-data era, 
sometimes representing more than 20\% of the budget in Data Centers nowadays. 
For an illustrative example, costs have exceeded 5 billion dollars per year 
over the last decade only in the US \cite{Benedict}, and it is predicted that 
the energy billing will increase in forthcoming years if power optimizations 
are not conducted in all levels, including operating systems, kernels and applications.

The industry is aware about the need of low-power CNN acceleration when using them 
extensively. Google is a clear example with Tensor Processing Unit tailored to 
their TensorFlow framework in its data centers, claiming that they are able 
to reduce power an order of magnitude versus GPUs \cite{TensorFlow16}.

The research community is also helping to reduce power on CNNs. Five notable examples 
recently published in 2016-17 are the following:

\begin{itemize}
    \item Moons \etal \cite{Moons16} propose methods at system and circuit level 
    based on approximate computing. They always perform training using 32-bit, 
    lowering precision during the test phase. They claim energy gains up to 30× 
    without losing classification accuracy and more than 100× at 99\% classification 
    accuracy, compared to a commonly used 16-bit fixed point number format.
    
    \item Cai \etal propose NeuralPower \cite{NeuralPower17}, a layer-wise predictive 
    framework based on sparse polynomial regression, for predicting the serving 
    energy consumption of a CNN deployed on different GPU platforms and Deep Learning 
    software tools, attaining an average accuracy of 88.24\% in execution time, 88.34\% 
    in power, and 97.21\% in energy.
    
    \item Andri \etal introduce YodaNN \cite{YodaNN16}, an energy and area efficiency 
    accelerator based on ASIC hardware optimized for BinaryConnect CNNs which basically 
    removes the need for expensive multiplications during training, also reducing I/O 
    bandwidth and storage.
    
    \item Yang \etal \cite{Yang17} propose an energy-aware pruning algorithm for CNNs 
    that directly uses energy consumption estimation of a CNN to guide the pruning process. 
    The energy consumption of AlexNet and GoogLeNet are reduced by 3.7x and 1.6x, respectively, 
    with less than 1\% top-5 accuracy loss. Results are obtained via a energy 
    estimation tool for Deep Neural Networks publicly available in \cite{CNN-energyestim}.
    
    \item Lin et al. \cite{Lin17} propose PredictiveNet to skip a large fraction of 
    convolutions in CNNs at runtime without modifying the CNN structure or requiring 
    additional branch networks. An analysis supported by simulations is provided to justify 
    how to preserve the mean square error (MSE) of the nonlinear layer outputs. 
    Energy savings are attained by reducing the computational cost by a factor of 2.9× 
    compared to a state-of-the-art CNN, while incurring marginal accuracy degradation.
    
\end{itemize}

Moving away from estimators, predictors and simulators, we may find examples of real 
energy measurements and studies on low-power devices like DSPs \cite{Mathew18} 
and FPGAs \cite{Bettoni17}, even for CNN applications \cite{Wang16,Mathew17}. 
But to the best of our knowledge, our work is pioneer on measuring the actual power 
consumption of CNNs with wires and measurement devices physically plugged to the pinout 
of latest GPU generations and multi-GPU platforms, and even identifying the most 
expensive operators and functions in terms of energy budget.


\section{CNN Overview}\label{s:CNN-overview}

Convolutional Neural Networks (CNNs) are a type of neural network particularly successful 
on computer vision problems where the information is spatially related and 
it can be represented in a hierarchical mode \cite{bengio2015book}. 
A CNN is defined by its architecture which is a set several convolutional layers 
and several fully connected layers. Each convolutional layer is, in general, 
the composition of a non-linear (convolutional filter) layer and a pooling 
or sub-sampling layer to get some spatial invariance.

In the last years, CNN models are standing out above on a wide range of applications, 
like object detection, text classification, natural language processing or scene labeling~\cite{krizhevsky2012nips,farabet2013,zhang2015,Wang2017cvpr}. 
CNNs are specially successful on data where the target can be represented with a feature 
hierarchy of increasing semantic complexity. When successfully trained, the output 
of the last hidden layer can be seen as the representation of the target in a high-level space. 
The fully connected layers reduce the dimensionality of the representation and hold 
the high-level knowledge, improving the classification accuracy. 

During training, a random batch, which is a set of $N$ samples, is selected from 
the training samples and passed through the model obtaining the activations (outputs) 
of each layer and the final output. This final output, depending on the type of application, 
can be a probability distribution (classification), an image (segmentation), 
a number (regression), etc. With this final output and its corresponding ground-truth label, 
a loss function designed for each problem computes the error, which is back-propagated 
from the top layers to the bottom ones. 

Therefore, during training, there are three different processes in a CNN: 

\begin{enumerate}
    \item Forward, where a batch is passed through the CNN to obtain the activations.
    \item Backward, where the error is back-propagated to obtain the derivatives.
    \item Weights update, for the weights of the model to be updated according to the solver. 
    This stage is negligible compared to the previous two and we have preferred to discard it 
    for the sake of simplicity. 
\end{enumerate}

During test, only the forward process is executed to obtain the final output of the model. 
Along with the back-propagation process, each layer computes its own derivatives according 
to the error coming from the top layer. Once derivatives are computed, the average derivative 
from the $N$ samples is computed and the weights of the model are updated according 
to the solver selected, with the Stochastic Gradient Descent (SGD) being the most common case. 

These three steps are repeated for $M$ epochs until the algorithm converges, 
with an epoch being a set of $M_b$ batches (or iterations) to process 
the whole training set. For example, in a training set composed of 1000 samples 
and batches of 100 samples, an epoch would have 10 batches or iterations.

This work focuses on energy consumption and execution time of the forward 
and backward processes, also analyzing the global accuracy for the model. 
Energy, acceleration and precision are put in perspective on modern GPUs 
as attractive candidates for a leadership on different models and problems, 
among which we select a bunch of popular instances for a representative case study.


\section{Our CNN Selection for Power Analysis}\label{s:CNN-selection}

We select two popular CNN architectures typically applied to process input data 
in computer vision, either using images or videos. On the image side, we deal 
with image recognition, that is, identify what appears in an image; whereas 
using videos we focus on gait, that is, the challenge of identifying people 
by the way they walk. We pretend this way to explore setups acting as solid 
templates for deep learning in computer vision, so that conclusions can easily 
be extrapolated to a wide range of problems.

The energy consumed by an algorithm is directly proportional to the number 
of operations and its type. In a CNN, this type is defined by the architecture 
and the kind of layers. The architecture also plays an important role on the 
number of operations, because that number increases with the number of layers. 
In addition, during training, more than one sample is passed through the 
CNN according to the mini-batch training process, and so the number of samples 
(batch size) influences power consumption and the convergence process 
in a decisive manner.

Each layer has its own number and type of operations, so we now characterize 
the most common layers used in the majority of CNNs.
For simplicity, all formulas are related to a single sample as input. 
Thus, when dealing with a batch, expressions must be multiplied by the batch size. 

We start introducing some terminology:
\begin{itemize}
    \item $w_{in}, h_{in}$. Width and height of the input sample, respectively.
    \item $w_{out}, h_{out}$. Width and height of the output sample, respectively.
    \item $ch_{in}, ch_{out}$. Number of input and output channels, respectively.
    \item $k_w, k_h$. Kernel width and height, respectively.
\end{itemize}
The terms $w_{out}$ and $h_{out}$ are obtained from the formula 
$w_{out} = \frac{(w_{in} - k_w) + 2P}{S}$, with $P$ being the padding 
applied to the input and $S$ the stride or step of the kernel. Similarly, 
$h_{out} = \frac{(h_{in} - k_h) + 2P}{S}$.

Using previous definitions, the functionality and number of operations performed 
at each layer are shown as follows. 

\myparagraph{Convolution. } It applies a kernel to the input sample. 
The number of operations is defined on Eq.~\ref{eq:conv}. In this layer, the type 
of operation is \textit{multiply\textendash accumulate (macc)}. In addition, 
if the convolution has bias, we need to include $ch_{out}$ \textit{add} operations.
\begin{equation}
    \#operations = (k_w \cdot k_h)(w_{out} \cdot h_{out}) \cdot ch_{in} \cdot ch_{out}
\label{eq:conv}
\end{equation}

\myparagraph{Fully connected. } It has full connections to all activations 
in the previous layer. The number of operations is defined on Eq.~\ref{eq:fc}. 
In this case, the type of operation is \textit{multiply\textendash accumulate (macc)}.
\begin{equation}
    \#operations = (w_{in} \cdot h_{in}) \cdot ch_{in} \cdot ch_{out}
\label{eq:fc}
\end{equation}

\myparagraph{Pooling. } It reduces the spatial size of the input to lower 
the amount of parameters and computation in the network. The number of operations 
is defined on Eq.~\ref{eq:pool}. The type of operation depends on the architecture, 
being \textit{max} the most common one.
\begin{equation}
    \#operations = (k_w \cdot k_h)(w_{out} \cdot h_{out}) \cdot ch_{in}
\label{eq:pool}
\end{equation}

\myparagraph{ReLU. } It applies a regularization function to the input. 
The number of operations is defined on Eq.~\ref{eq:relu}. In this case, 
the type of operation is \textit{max}.
\begin{equation}
    \#operations = (w_{in} \cdot h_{in}) \cdot ch_{in}
\label{eq:relu}
\end{equation}

\myparagraph{Dropout. } It randomly disconnects inputs to minimize overfitting. 
The number of operations is defined on Eq.~\ref{eq:dropout}. In this case, 
the type of operation is \textit{multiplication} by $0$ or $1$ depending on 
the the input to be disconnected or not.
\begin{equation}
    \#operations = (w_{in} \cdot h_{in}) \cdot ch_{in}
\label{eq:dropout}
\end{equation}

\myparagraph{Batch normalization. } It normalizes the input subtracting 
the mean and dividing by the standard deviation. The number of operations 
is defined on Eq.~\ref{eq:bn}. The type of operations are \textit{add} and 
\textit{division}. As the number of both operations is the same, we combine 
them into a single equation.

\begin{equation}
    \#operations = (w_{in} \cdot h_{in}) \cdot ch_{in} \cdot 2
\label{eq:bn}
\end{equation}

\myparagraph{Softmax. } It scales a $K-$dimensional vector of arbitrary real values 
to a $K-$dimensional vector of real values in the range $[0, 1]$ that add up to $1$. 
The number of operations is defined on Eq.~\ref{eq:softmax}. The type of operations 
are \textit{exponential}, \textit{add} and \textit{division}. As the number of the 
three operations is the same, we combine them into a single equation.
\begin{equation}
    \#operations = (w_{in} \cdot h_{in}) \cdot ch_{in} \cdot 3
\label{eq:softmax}
\end{equation}

Apart from the number and type of operations, each layer is also characterized 
by the data volume read and written. In this analysis, the formulas are valid 
for all layers so we present a formula for the reading process~\ref{eq:reading} 
and another one for the writing one~\ref{eq:writing}. Note that in the reading part, 
the second term refers to the weights of the layer.
\begin{equation}
    read = (w_{in} \cdot h_{in}) \cdot ch_{in} + (k_w \cdot k_h) \cdot ch_{in}
\label{eq:reading}
\end{equation}
\begin{equation}
    written = (w_{out} \cdot h_{out}) \cdot ch_{out}
\label{eq:writing}
\end{equation}

For building our benchmark, we first select four architectures:
(1) a 2D-CNN~\cite{castro2017iwann} based on AlexNet~\cite{krizhevsky2012nips} 
using videos as inputs, (2) a ResNet network~\cite{he2016cvpr} specifically developed 
for gait recognition~\cite{castro2017biosig} involving videos, (3) CaffeNet, 
which is an implementation of AlexNet released with Caffe, and (4) ResNetIm,
which is the ResNet34 published in~\cite{he2016cvpr}. Then, for each architecture, 
we select three batch sizes so that we can characterize the energy consumed 
and accuracy depending on networks and batch sizes.

Among those hyper-parameters to be optimized within a DL network, we have selected 
the one which has a bigger impact in performance and accuracy. Other candidates 
might be the learning rate, to affect the convergence speed, and the stride 
of the convolutions, which defines the step size of the convolutions applied 
to an image. The learning rate only affects if a good value is known beforehand 
to guarantee a fast convergence, but in most cases that value is a heuristic 
determined through an exhaustive experimental process. The stride has a huge impact 
in the performance and energy (less operations are performed with higher stride values), 
and also in the model, because networks can lose important local information. 
Nevertheless, latest networks like ResNet just use stride one and small convolutions 
to capture that information, leaving this hyper-parameter with a minor influence 
and highly sensitive to the problem itself.


\subsection{Architecture analysis}

\begin{figure}[t]   
\begin{center}
   \includegraphics[width=0.8\linewidth]{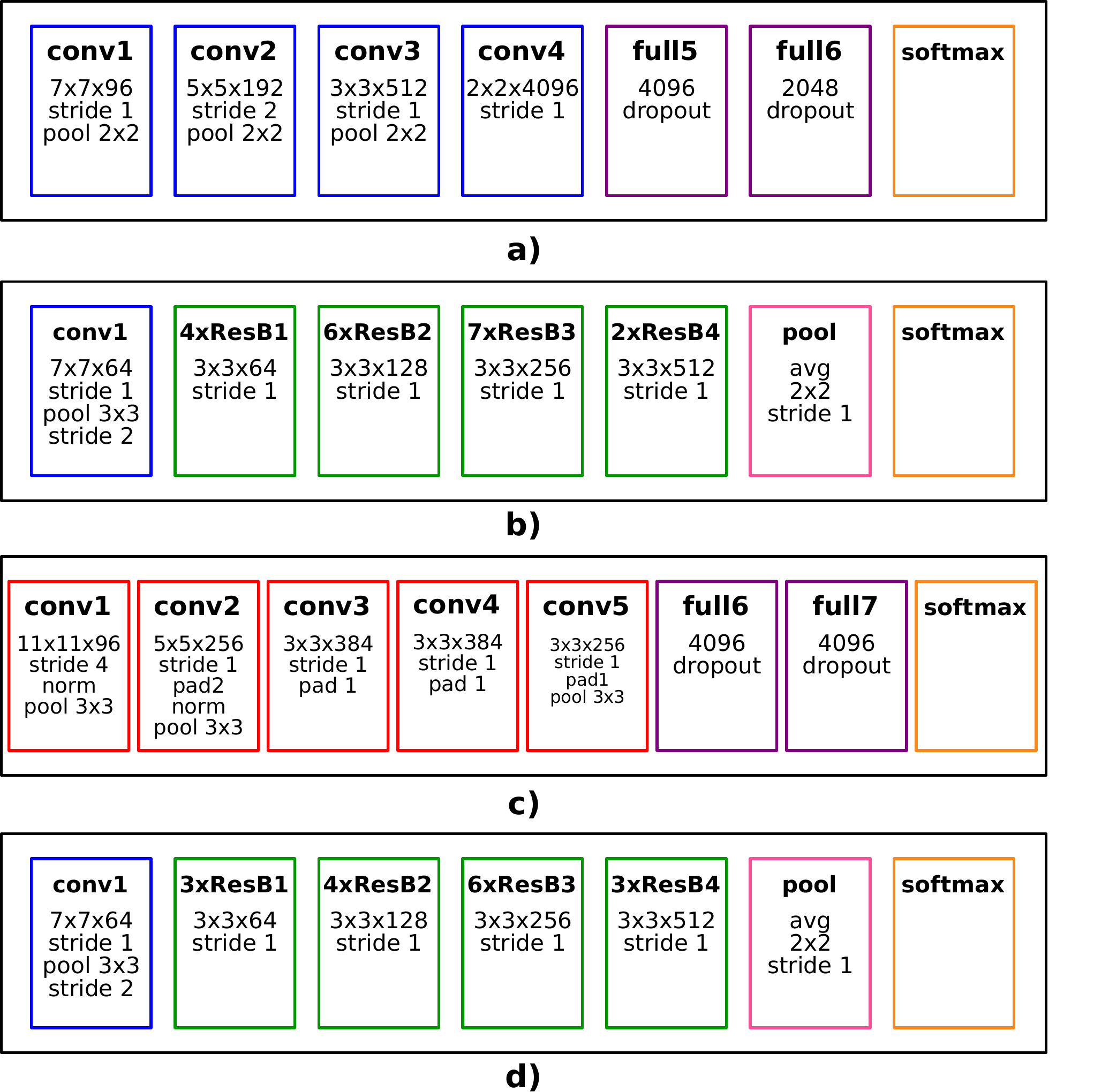}
\end{center}
\caption{Proposed CNN models for gait signature extraction. a) 2D-CNN: linear 
CNN with four 2D convolutions, two fully connected layers and a softmax classifier. 
b) ResNet: residual CNN with a 2D convolution, four residual blocks, 
an average pooling layer and a final softmax classifier. Note that before 
the first block of each kind (ResB 1, 2, 3, 4), there is an adapter convolution 
to resize the input image to that of the next block. c) CaffeNet: linear CNN with 
five 2D convolutions, two fully connected layers and a softmax classifier. 
d) ResNetIm: residual CNN with a 2D convolution, four residual blocks, 
an average pooling layer and a final softmax classifier. Note that before 
the first block of each kind (ResB 1, 2, 3, 4), there is an adapter convolution 
to resize the input image to that of the next block.}
\label{fig:archCNN}
\end{figure}

\myparagraph{2D-CNN (15 layers).}

This architecture is inspired by AlexNet~\cite{krizhevsky2012nips} 
and it is adapted to the specific requirements of gait recognition. 
For our particular case, we use optical flow as input following the 
approach described in~\cite{castro2017iwann}.

The proposed CNN comprises the following sequence of layers (see Fig.~\ref{fig:archCNN}.a):

\begin{enumerate}
  \item `\textit{conv1}', 96 filters of size $7\times 7$ applied with stride 1 
  followed by max pooling $2\times 2$.
  \item `\textit{conv2}', 192 filters of size $5\times 5$ applied with stride 2 
  followed by max pooling $2\times 2$.
  \item `\textit{conv3}', 512 filters of size $3\times 3$ applied with stride 1 
  followed by max pooling $2\times2$.
  \item `\textit{conv4}', 4096 filters of size $2\times 2$ applied with stride 1.
  \item `\textit{full5}', fully-connected layer with 4096 units and dropout.
  \item `\textit{full6}', fully-connected layer with 2048 units and dropout.
  \item `\textit{softmax}', softmax layer with as many units as subject identities.
\end{enumerate}

\myparagraph{ResNet (167 layers).}

This CNN is composed of a sequence of layers and residual blocks shown 
in Fig.~\ref{fig:archCNN}.b
(two consecutive convolutions of size $3 \times 3$ plus a sum layer, 
as defined in \cite{he2016cvpr}). The main blocks in our model are: 

\begin{enumerate}
  \item `\textit{conv1}', 64 filters of size $7\times 7$ applied with stride 1 
  followed by max pooling $3\times 3$ with stride 2.
  \item `\textit{4xResB1}', 4 residual blocks with convolutions of 64 filters 
  of size $3\times 3$ applied with stride 1.
  \item `\textit{6xResB2}', 6 residual blocks with convolutions of 128 filters 
  of size $3\times 3$ applied with stride 1.
  \item `\textit{7xResB3}', 7 residual blocks with convolutions of 256 filters 
  of size $3\times 3$ applied with stride 1.
  \item `\textit{2xResB4}', 2 residual blocks with convolutions of 512 filters 
  of size $3\times 3$ applied with stride 1.
  \item `\textit{pool}', global average pooling $2\times2$.
  \item `\textit{softmax}', softmax layer with as many units as subject identities.
\end{enumerate}

\myparagraph{CaffeNet (25 layers).}

This architecture is inspired by AlexNet~\cite{krizhevsky2012nips}, 
and it was released within Caffe, to be used for object recognition in images. 
The input has a size of $227\times 227$, obtained from a random crop of the original 
images resized to $256\times 256$.

The proposed CNN is composed by the following sequence of layers (see Fig.~\ref{fig:archCNN}.c):

\begin{enumerate}
  \item `\textit{conv1}', 96 filters of size $11\times 11$ applied with stride 4 
  followed by max pooling $3\times 3$ and a normalization layer of size $5\times 5$.
  \item `\textit{conv2}', 256 filters of size $5\times 5$ applied with stride 1 
  and padding 2 followed by max pooling $3\times 3$ and a normalization layer of size $5\times 5$.
  \item `\textit{conv3}', 384 filters of size $3\times 3$ applied with stride 1 and padding 1.
  \item `\textit{conv4}', 384 filters of size $3\times 3$ applied with stride 1 and padding 1.
  \item `\textit{conv5}', 256 filters of size $3\times 3$ applied with stride 1 and padding 1 
  followed by max pooling $3\times 3$.
  \item `\textit{full6}', fully-connected layer with 4096 units and dropout.
  \item `\textit{full7}', fully-connected layer with 4096 units and dropout.
  \item `\textit{softmax}', softmax layer with as many units as object identities.
\end{enumerate}

\myparagraph{ResNetIm (94 layers).}

This CNN is composed of a sequence of layers and residual blocks shown 
in Fig.~\ref{fig:archCNN}.d
(two consecutive convolutions of size $3 \times 3$ plus a sum layer, 
as defined in \cite{he2016cvpr}). The main blocks in our model are: 

\begin{enumerate}
  \item `\textit{conv1}', 64 filters of size $7\times 7$ applied with stride 1 
  followed by max pooling $3\times 3$ with stride 2.
  \item `\textit{3xResB1}', 3 residual blocks with convolutions of 64 filters 
  of size $3\times 3$ applied with stride 1.
  \item `\textit{4xResB2}', 4 residual blocks with convolutions of 128 filters 
  of size $3\times 3$ applied with stride 1.
  \item `\textit{6xResB3}', 6 residual blocks with convolutions of 256 filters 
  of size $3\times 3$ applied with stride 1.
  \item `\textit{3xResB4}', 3 residual blocks with convolutions of 512 filters 
  of size $3\times 3$ applied with stride 1.
  \item `\textit{pool}', global average pooling $2\times2$.
  \item `\textit{softmax}', softmax layer with as many units as subject identities.
\end{enumerate}

The convolutional layers from all CNNs use the rectification (ReLU) activation function.
Applying the formulas of Section \ref{s:CNN-selection} to characterize our networks, 
we may obtain the number of arithmetic operations and memory accesses required, 
which are compiled in Table \ref{t:operations}. In addition, we show the Computation 
to Communication ratio~\cite{zhang2015ctc} defined as: $CTC = \frac{N_{op}}{N_d}$ 
where $N_{op}$ is the total number of operations and $N_d$ is the amount of data read.

\begin{table}   
\caption{Characterization of our CNN networks through number of arithmetic operations, 
data volume read and written per sample during forward step and ratio between 
the number of operations and amount of data read.}\label{t:operations}
\centering
\tabsize
\begin{tabular}{l|c|cc|c}
\hline 
Type of CNN & \# arithmetic operations & Data volume read & Data volume written & CTC ratio \\ 
\hline 
ResNet      &       425.13M            &     87.6 MB      &      4.8 MB         &     18.5  \\ 
\hline 
2D-CNN      &       783.84M            &     73.8 MB      &      1.9 MB         &     40.5  \\ 
\hline 
CaffeNet    &       727.20M            &    239.1 MB      &      6 MB           &     11.6  \\ 
\hline 
ResNetIm    &       2080.96M           &     97.1 MB      &     46.6 MB         &     81.8  \\ 
\hline 
\end{tabular} 
\end{table}

\subsection{Batch analysis}\label{s:batch}

The batch size (number of samples) used during training influences three aspects of the model: 
number of operations, performance and accuracy. More precisely, the number of operations 
is defined by the Eq.~\ref{eq:operations_batch} and the data read and written by 
Eq.~\ref{eq:read_batch} and~\ref{eq:written_batch} respectively.

\begin{equation}
    \#operations = B \cdot operationsSample
\label{eq:operations_batch}
\end{equation}

\begin{equation}
    readB = B \cdot readSample
\label{eq:read_batch}
\end{equation}

\begin{equation}
    writtenB = B \cdot writtenSample
\label{eq:written_batch}
\end{equation}

where $B$ represents the batch size, $operationsSample$ the number of operations, 
$readSample$  the data read and $writtenSample$ the data written, obtained by the 
formulas described in Section \ref{s:CNN-selection} for one sample.

The batch size defines the number of samples used as input to a model. Therefore, 
the bigger the batch size, the more number of operations performed as there are 
more samples to process. If we consider the latency due to input data coming from 
secondary memory, a bigger batch size allows a better overlapping between computations 
on GPU and CPU to GPU communications. Moreover, we have to remember that the batch size 
plays an important role during training as the weights are updated according to the mean 
of the gradients obtained from the images of the batch. Therefore, there is a trade 
off here: bigger batches improve accuracy in gradients, but smaller batches (noisy 
gradients) benefit convergence as it maximizes the exploration of the solution space. 
Taking into account all these considerations, we are going to evaluate three 
batch sizes: 64, 128 and 256. These values are the most common in the literature 
and they achieve good results in terms of accuracy for the problems tested here, 
and constitute the best candidates in our quest for the optimal batch size 
in terms of accuracy, performance and power requirements.

\subsection{Energy measurement}

The training process is composed of three main parts, namely, forward, backward and 
weight updating (see Section \ref{s:CNN-overview}). According to our experiments, 
forward and backward steps consume on average more than $95\%$ of the execution time. 
Then, we focus on the two first steps to simplify our analysis. In any case, 
total values can always be roughly obtained by adding this percentage to the sum 
of forward and backward steps.

Algorithm~\ref{alg:1} shows the training process and the points established 
for measurements. Since the algorithm executes concurrently on the GPU, 
we use CUDA events to make sure that the measured execution is over. 
Also, our infrastructure for measuring power is attached to a single GPU, 
which means that on multi-GPU executions, we nominate a root GPU and 
the global consumption is extrapolated for the set involved. 
More details are provided later in Section \ref{s:GPU-impl}. 
Note that $maxIter$ is computed as the number of iterations in an epoch 
multiplied by the number of epochs.

\begin{algorithm}[htbp]
\begin{algorithmic}
\STATE model = InitializeModel()
\STATE iter = 0
\STATE maxIter = Initialize(InitParams)
\WHILE{iter $<$ maxIter} 
    \STATE StartTimer()
	\STATE data = LoadData(batch)
	\STATE time = StopTimer()
	\STATE StartTimerAndPowerMeassurement()
	\STATE output = Forward(model, data)
	\STATE time, power = StopTimerAndPowerMeassurement()
	\STATE StartTimerAndPowerMeassurement()
	\STATE derivatives = Backward(model, output)
	\STATE time, power = StopTimerAndPowerMeassurement()
	\STATE model = updateWeights(model, derivatives)
	\STATE iter++
\ENDWHILE	
\end{algorithmic}
\caption{Schematic for CNN training.}
\label{alg:1}
\end{algorithm}


\section{GPU Implementation}\label{s:GPU-impl}

We use Caffe~\cite{jia2014caffe} (commit c98de53b7817c732b482c2fa810f09c260c58857) 
with cuDNN~\cite{chetlur2014cudnn} 6.0, NCCL 2.1.2 and CUDA 8.0 libraries 
to train our CNNs. Forward and backward processes are entirely implemented in GPU 
by Caffe using the primitives available in cuDNN. To update the weights 
efficiently in a multi-GPU environment, Caffe uses primitives included in NCCL. 
Finally, the CPU just loads the input data.

When the model is being trained on a single GPU, we use a CPU thread to load data 
constantly from secondary storage into a CPU memory buffer. This way, we maximize 
overlapping between data transfers and GPU computation. If there is enough data 
to fill a batch, the GPU computes the forward and backward steps while the CPU 
is loading new data. Finally, the weights are updated in the GPU and the process 
starts again to compute a new batch.

When the model is being trained using multiple GPUs, each GPU has a CPU thread 
which is loading data into its own memory buffer, thus, we have one thread and 
one memory buffer per GPU. In this case, each GPU has exactly the same 
CNN architecture model with similar weights, but the batch is divided among GPUs 
(e.g. a batch with 64 samples trained with 2 GPUs is splitted into 2 sub-batches 
of 32 samples). Once  all GPUs have finished their computation, the derivatives 
are collected and the weights updating process starts. In order to optimize 
the weights updating step, Caffe uses the \textsf{ncclAllReduce}
\footnote{https://images.nvidia.com/events/sc15/pdfs/NCCL-Woolley.pdf} primitive, 
which gathers local gradients, computes global derivatives and leaves a copy of 
those on each GPU. It is important to clarify that according to the documentation, 
Caffe uses a tree reduction strategy 
\footnote{https://github.com/BVLC/caffe/blob/master/docs/multigpu.md}, 
but recent implementations use NCCL instead to improve the multi-GPU performance. 
Thus, \textsf{ncclAllReduce} arranges the GPUs in a virtual ring and the information 
to be transferred is split into small packages. Then, the $i$-th GPU transfers 
a package to its neighbour ($i+1$) and, at the same time, performs the reduction 
computation with the information coming from the $(i-1)$-th GPU. This process 
is repeated until all packages are transferred and all reductions are done. 
When the primitive ends, all GPUs store the same information, that is, the values 
of global derivatives. That way, each GPU updates the model independently.


\section{Monitoring Energy}\label{s:monitoring-energy}

\subsection{Measurement Infrastructure}

\begin{figure}   
\centering
\includegraphics[width=\textwidth]{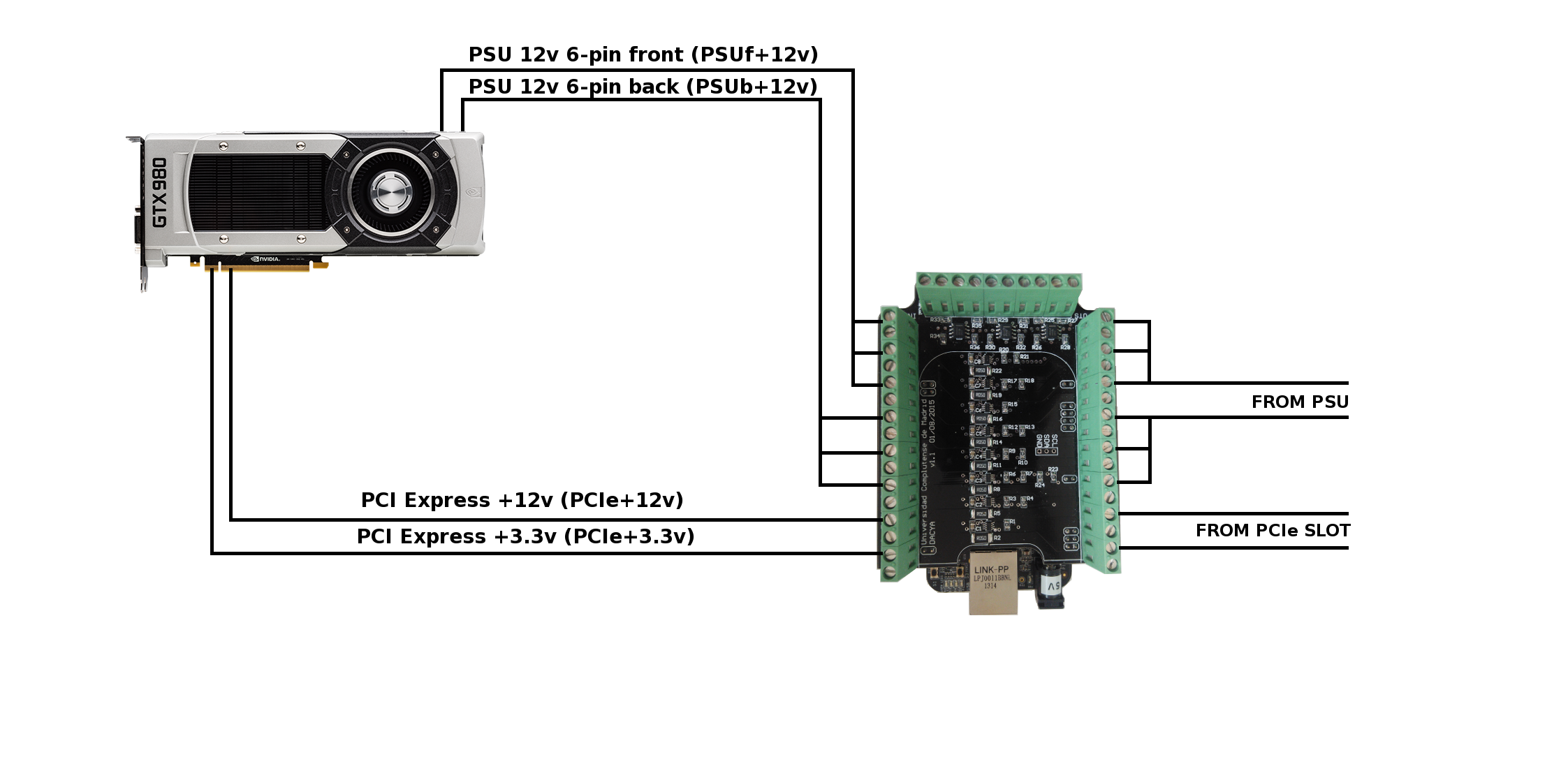}
\vspace*{-2cm}
\caption{Wires, slots, cables and connectors for measuring energy on GPUs.}
\label{f:measure-GPU}
\end{figure}

We have built a system to measure current, voltage and wattage based on 
a Beaglebone Black, an open-source hardware \cite{BeagleBoard} combined with 
the Accelpower module \cite{AccelPowerCape}, 
which has eight INA219 sensors \cite{Adafruit15}. Inspired by \cite{Igual15}, 
wires taken into account are two power pins on the PCI-express slot (12 and 3.3 volts) 
plus six external 12 volt pins coming from the power supply unit (PSU) in the form of 
two supplementary 6-pin connectors (half of the pins used for grounding). 
See Figure \ref{f:measure-GPU} for details.

\subsection{Software tool}

Accelpower uses a modified version of {\tt pmlib} library \cite{Alonso12}, 
a software package specifically created for monitoring energy.
It consists of a server daemon that collects power data from devices 
and sends them to the clients, together with a client library for communication 
and synchronization with the server.

\subsection{Methodology for Measuring Energy}

\begin{figure}  
\centering
\includegraphics[width=0.8\textwidth]{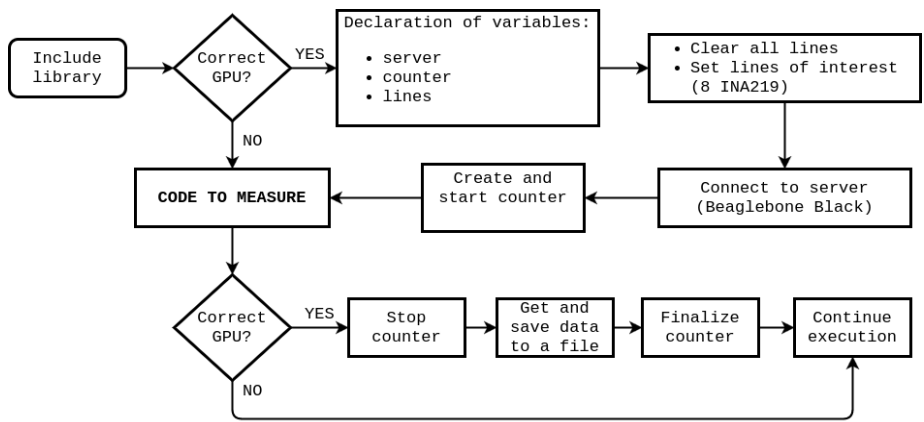}
\caption{Flow diagram for measuring energy on a code excerpt when running on the GPU.}
\label{f:measure-flow}
\end{figure}

The methodology for measuring energy begins with a start-up of the server daemon. 
Then, the source code of the application where the energy wants to be measured 
has to be modified to (1) declare {\tt pmlib} variables, (2) clear and set the wires 
which are connected to the server, (3) create a counter and (4) start it. 
Once the code is over, we (5) stop the counter, (6) get the data,
(7) save them to a .csv file, and (8) finalize the counter. 
See Figure \ref{f:measure-flow} for a flow chart. Note that we get the 
instant consumption per measurement. Therefore, to obtain the global consumption, 
we compute the discrete integral over time.

Note that the homemade system we have built to measure real energy consumption
can only be attached to a single GPU. That way, for obtaining performance or accuracy
numbers, a single run suffices in multi-GPU environments, but energy requires a different
approach. We have to move our measurement infrastructure from Pascal GPU 
to Maxwell GPU and perform different runs to monitor power values in both architectures. 
The final values are obtained multiplying those previous values by the number of GPUs. 
In case of four GPUs, we compute the energy for Maxwell and Pascal and  aggregate them.

\subsection{Hardware Resources}

Our experimental study was conducted on a multi-GPU computer endowed with 
an Intel Xeon E5-2620 server and four PCI 3.0 slots to hold up to two 
Nvidia Titan X Pascal and two Titan X Maxwell GPUs.
Table \ref{t:HW} summarizes major features for those GPUs. Note that cores 
and memory frequencies are overclocked to the maximum allowed by each GPU.
The CPU has eight cores running at 2100 MHz and 64 GB of main memory running 
at 2400 MHz in a four-channel architecture. For secondary storage, we enable 
a Samsung SSD 850 EVO with a sequential reading up to 540 MB/s and an access 
time of 0.03 ms. On the software side, Ubuntu 14.04.4 LTS 64 bits was installed 
as the operating system together with CUDA 8.0.

\begin{table}   
\caption{The set of Nvidia GPUs used along our experimental study.}\label{t:HW}
\centering
\tabsize
\begin{tabular}{l|c|c}
\toprule
Commercial model                 & {\bf Titan X} & {\bf Titan X}  \\ 
GPU generation and year          &  Maxwell 2015 &   Pascal 2016  \\ \midrule
\multicolumn{2}{l}{\bf Raw computational power:} \\ \midrule
Number of cores                  &     3072      &      3584      \\
Cores frequency                  &    1392 MHz   &    1911 MHz    \\
Peak processing                  &   6.6 TFLOPS  &    11 TFLOPS   \\
CUDA Compute Capability          &      5.2      &      6.1       \\ \midrule
\multicolumn{2}{l}{\bf Dynamic memory (DRAM):} \\ \midrule
Size and type                    &  12 GB GDDR5X & 12 GB GDDR5X   \\
Frequency and width              & 3505 MHz @ 384 bits & 5005 MHz @ 384 bits \\
Bandwidth                        &   336.5 GB/s  &   480 GB/s.    \\ \midrule
\multicolumn{2}{l}{\bf Static memory (cache):} \\ \midrule
Shared memory per multiprocessor &   48 Kbytes   &   48 Kbytes    \\
L2 cache                         &    3 Mbytes   &    3 Mbytes    \\ \midrule
\multicolumn{2}{l}{\bf Thermal and energy specifications:} \\ \midrule
Maximum GPU Temperature          &     91 C      &      94 C      \\
Peak Power Consumption (TDP)     &     250 W     &     250 W      \\ 
Recommended supply Power         &     600 W     &     600 W      \\ \bottomrule
\end{tabular}
\end{table}


\section{Input datasets}\label{s:datasets}

We cover a quantitative and qualitative analysis for a multi-GPU system, 
where each GPU executes evenly a subset or partition of the computation 
according to the workload distribution.
Our experiments are conducted on a challenging dataset for gait recognition, 
TUM-GAID~\cite{hofmann2014tumgaid}, and a huge dataset for image recognition, 
ILSVRC12~\cite{ILSVRC15}.

\subsection{TUM-GAID}

\begin{figure}[tb]   
\begin{center}
   \includegraphics[width=0.99\linewidth]{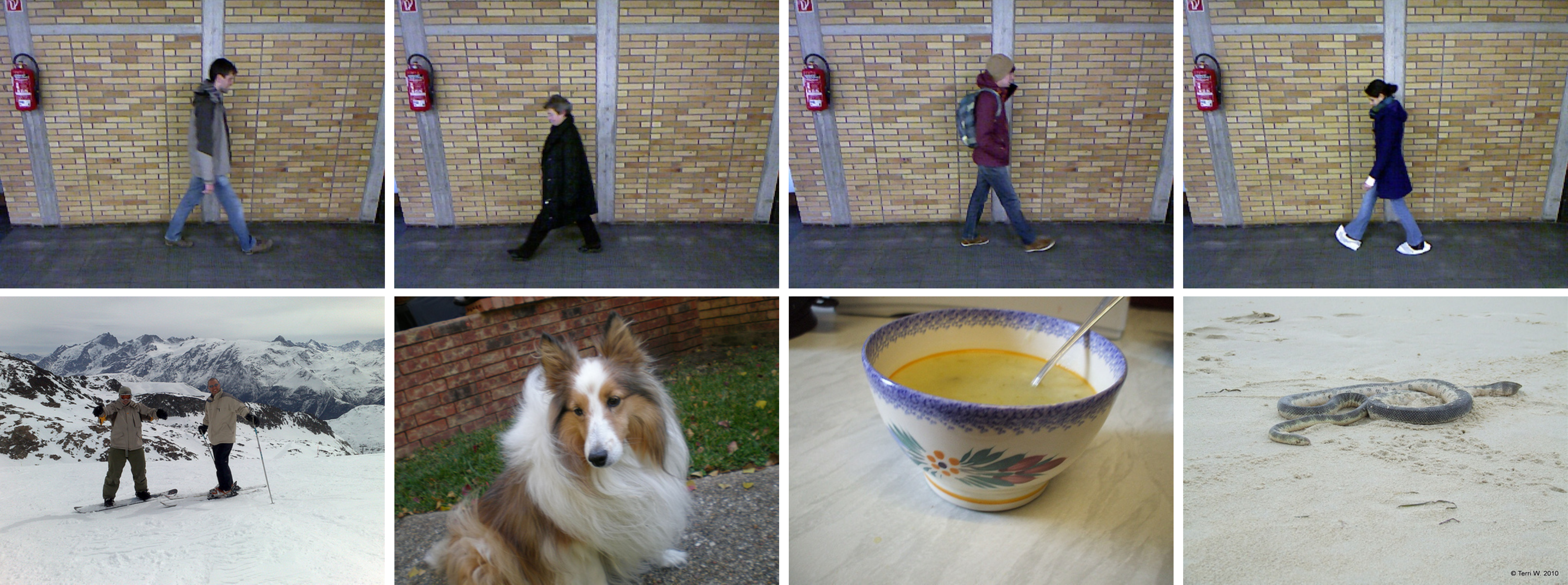}
\end{center}
\caption{Our input datasets. Upper: TUM-GAID images from different subjects. 
Lower: ILSVRC1 images from different classes.
}
\label{fig:datasets}
\end{figure}

TUM-GAID (TUM Gait from Audio, Image and Depth) collects 305 subjects performing 
two walking trajectories in an indoor environment. The first trajectory is traversed 
from left to right and the second one from right to left. Two recording sessions 
were performed, one in January, where subjects wore heavy jackets and mostly 
winter boots, and another one in April, where subjects wore lighter clothes. 
The action is captured by a Microsoft Kinect sensor which provides a video stream 
with a resolution of $640 \times 480$ pixels and a frame rate around 30 FPS. 
Figure ~\ref{fig:datasets} provides some examples.

Hereinafter the following nomenclature is used to refer each of the four walking 
conditions considered: \textit{normal} walk (\textit{N}), carrying a \textit{backpack} 
of approximately 5 kg (\textit{B}), wearing coating \textit{shoes} (\textit{S}, 
as used in clean rooms for hygiene conditions), and \textit{elapsed time} (\textit{TN-TB-TS}).
During our experiments, we follow the experimental protocol defined by the authors 
of the dataset \cite{hofmann2014tumgaid}.

\subsection{ILSVRC12}

ImageNet Large-Scale Visual Recognition Challenge 2012 (ILSVRC12) is an annual 
competition which uses a subset of ImageNet. This subset is composed of 1000 classes 
with more than 1000 images per class. In total, there are roughly 1.2 million 
training images, 50,000 validation images, and 150,000 testing images. 
Those images have a variable resolution and have been manually annotated.

At test time, it is customary to report two accuracy rates: top-1 and top-5, 
where top-1 value is the classic accuracy metric and the top-5 accuracy rate 
is the fraction of test images for which the correct label is among the five 
most frequent labels considered by the model.

\subsection{Customizing videos and images}

For the experiments with ResNet and 2D-CNN, we resize all videos to a common 
resolution of $80 \times 60$ pixels, keeping the original aspect ratio of 
video frames. This size exhibits a good trade-off between computational cost and 
recognition performance, as already reflected in a previous work ~\cite{castro2017iwann}.

Given the resized video sequences, we compute dense \textit{OF} on pairs of frames 
by using the method of Farneback \cite{Farneback03} implemented in OpenCV 
library~\cite{opencv_library}. For each video frame, two \textit{OF} frames 
are generated containing the $x$ and $y$ components of the flow vector. 
In parallel, people are located in a rough manner along the video sequences 
by background subtraction~\cite{kaewtrakulpong2002bmm}. 
Then, we crop the video frames to remove part of the background and to align 
the subsequences (people are $x$-located in the middle of the central frame, \#13), 
obtaining video frames of $60\times 60$ pixels (keeping the whole height).

Finally, from the cropped \textit{OF} maps, we build subsequences of $25$ frames 
by stacking \textit{OF} maps with an overlap of $\mathcal{O}\%$ frames. In our case, 
we choose $\mathcal{O}=80\%$, that is, to build a new subsequence, we use $20$ frames 
of the previous subsequence and $5$ new frames. For most state-of-the-start datasets, 
25 frames cover almost one complete gait cycle~\cite{barnich2009prl}. Consequently, 
each \textit{OF} volume has a size of $60\times 60 \times 50$, which constitutes 
a sample for 2D-CNN and ResNet.  

To increase the amount of training samples, we add mirror sequences and apply 
spatial displacements of $\pm 5$ pixels on each axis, obtaining a total of 
8 new samples from each original sample.

For the experiments with CaffeNet, we resize all the images to a common resolution 
of $256 \times 256$ pixels and, like in previous works~\cite{krizhevsky2012nips}, 
we do not keep the aspect ratio. During training, we perform random cropping and 
mirroring to obtain samples of $227 \times 227$. In this case, we do not perform 
spatial displacements, but center cropping at test time.

During ResNet and 2D-CNN training, the weights are learnt using mini-batch 
stochastic gradient descent algorithm with momentum equal to $0.9$. We set 
weight decay to $5 \cdot 10^{-4}$ and dropout to $0.4$ (when required). 
The number of epochs is limited to 20 and the learning rate is initially 
set to $10^{-2}$, to be decreased a $20\%$ every epoch. 

During CaffeNet training, the weights are also learnt using mini-batch 
stochastic gradient descent algorithm with momentum equal to $0.9$. 
We set weight decay to $5 \cdot 10^{-4}$ and dropout to $0.5$. 
The number of epochs is limited to 90 and the learning rate is initially 
set to $10^{-2}$, to be divided by $10$ every 20 epochs.


\section{Experimental Results}\label{s:exper-results}

Our testbed was executed on a multi-GPU server endowed with two Titan X Pascal 
and two Titan X Maxwell GPUs. The infrastructure for measuring time and energy 
was migrated from a Pascal GPU to a Maxwell one to gather all results shown 
along this section. For the sake of reliability and variance, we run all 
our experiments three times, and take the average as valid number. By using
the same seed for the three experiments, the training process matches in all cases.
Moreover, our tables distinguish rows in white for those CNNs using the 
TUM-GAID dataset (ResNet and 2D-CNN), and rows shaded for those CNNs using 
the ILSVRC12 dataset (CaffeNet and ResNetIm).

Table~\ref{t:iterations} shows the number of iterations and epochs executed for 
each CNN. Note that the number of epochs is the same along batch sizes 
but the number of iterations (i.e. batches) changes with the batch size. 
That way, the larger the batch size, the more samples are executed per iteration 
and less iterations are required to process training data.

Before we start discussing the experiments, let us introduce the section contents. 
In Sections~\ref{s:Pascal-results} and~\ref{s:maxwell-results}, execution time 
and energy consumption are measured and discussed for forward and backward steps 
at batch level on Pascal and Maxwell GPUs. 
Also, different metrics are employed to compare and choose the best experimental setup 
for each device. In Section~\ref{s:PascalvsMaxwell}, a comparison between 
GPU generations is performed in terms of execution time and energy consumption. 
Section~\ref{s:EnergyvsPerf} performs a similar comparison for energy versus performance. 
Section~\ref{s:accuracy} evaluates accuracy for the different trained models because 
a good model from either an energy or performance viewpoint is useless without 
a decent accuracy. Finally, Section~\ref{s:best-approach} provides a guideline 
to select the best settings according to our findings.

\begin{table}  
\caption{Iterations and epochs run for each CNN model and batch size.}\label{t:iterations}
\centering
\tabsize
\begin{tabular}{l|rr|rr|rr}
\hline 
& \multicolumn{2}{c|}{Batch size: 64} & \multicolumn{2}{c|}{Batch size: 128} & \multicolumn{2}{c}{Batch size: 256} \\
CNN model                    & Iterations & Epochs & Iterations & Epochs & Iterations & Epochs \\ \hline 
ResNet                       &     72 368 &     20 &     36 184 &     20 &     18 092 &     20 \\ 
2D-CNN                       &     72 368 &     20 &     36 184 &     20 &     18 092 &     20 \\ 
\rowcolor{gray!30} CaffeNet  &  1 800 000 &     90 &    900 000 &     90 &    450 000 &     90 \\ 
\rowcolor{gray!30} ResNetIm  &  1 800 000 &     90 &    900 000 &     90 &    450 000 &     90 \\ 
\hline
\end{tabular} 
\end{table}

\subsection{Results on Pascal}\label{s:Pascal-results}   

\begin{table}   
\caption{Execution times (in seconds) on a Pascal GPU for forward and backward steps 
of four CNN models using one, two and four GPUs and three batch sizes: 64, 128 and 256.}
\label{t:pascal}
\centering
\tabsize
\begin{tabular}{c|l|rrr|rrr|rrr}
\multicolumn{11}{c}{Forward} \\ \cline{3-11}
\multicolumn{2}{c|}{~} & \multicolumn{3}{c|}{Seconds per batch} & \multicolumn{3}{c|}{Samples per second} & \multicolumn{3}{c}{Seconds for whole training} \\ 
\multicolumn{2}{r|}{Batch size $\longrightarrow$} & 64 & 128 & 256 & 64 & 128 & 256 & 64 & 128 &   256 \\ \hline 
                   &   ResNet & 0.029 & 0.041 & 0.067 & 2 207 &  3 122 &  3 821 &   2 099 &   1 484 &  1 212 \\  
1 GPU              &   2D-CNN & 0.013 & 0.029 & 0.057 & 4 923 &  4 414 &  4 491 &     941 &   1 049 &  1 031 \\ 
\rowcolor{gray!30} & CaffeNet & 0.013 & 0.027 & 0.053 & 4 923 &  4 741 &  4 830 &  23 400 &  24 300 & 23 850 \\ 
\rowcolor{gray!30} & ResNetIm & 0.075 & 0.144 &   -   &   853 &    889 &   -    & 135 000 & 129 600 &    -   \\ \hline 
                   &   ResNet & 0.025 & 0.030 & 0.041 & 2 560 &  4 267 &  6 244 &   1 809 &   1 086 &    742 \\ 
2 GPUs             &   2D-CNN & 0.008 & 0.014 & 0.027 & 8 000 &  9 143 &  9 481 &     579 &     507 &    488 \\ 
\rowcolor{gray!30} & CaffeNet & 0.008 & 0.014 & 0.027 & 8 000 &  9 143 &  9 481 &  14 400 &  12 600 & 12 150 \\
\rowcolor{gray!30} & ResNetIm & 0.040 & 0.074 & 0.138 & 1 600 &  1 730 &  1 855 &  72 000 &  66 600 & 62 100 \\ \hline 
                   &   ResNet & 0.029 & 0.032 & 0.040 & 2 207 &  4 000 &  6 400 &   2 099 &   1 158 &    724 \\ 
4 GPUs             &   2D-CNN & 0.007 & 0.012 & 0.023 & 9 143 & 10 667 & 11 130 &     507 &     434 &    416 \\ 
\rowcolor{gray!30} & CaffeNet & 0.007 & 0.014 & 0.024 & 9 143 &  9 143 & 10 667 &  12 600 &  12 600 & 10 800 \\
\rowcolor{gray!30} & ResNetIm & 0.035 & 0.062 & 0.112 & 1 829 &  2 065 &   2286 &  63 000 &  55 800 & 50 400 \\ \hline
\multicolumn{11}{c}{~} \\
\multicolumn{11}{c}{Backward} \\ \cline{3-11}
\multicolumn{2}{c|}{~} & \multicolumn{3}{c|}{Seconds per batch} & \multicolumn{3}{c|}{Samples per second} & \multicolumn{3}{c}{Seconds for whole training} \\
\multicolumn{2}{r|}{Batch size $\longrightarrow$} & 64 & 128 & 256 &   64 & 128 & 256 & 64 & 128 & 256 \\ \hline 
                   &   ResNet & 0.125 & 0.220 & 0.254 &   512 &   582 & 1 008 &   9 046 &   7 960 &   4 595 \\  
1 GPU              &   2D-CNN & 0.021 & 0.048 & 0.094 & 3 048 & 2 667 & 2 723 &   1 520 &   1 737 &   1 701 \\ 
\rowcolor{gray!30} & CaffeNet & 0.026 & 0.053 & 0.105 & 2 462 & 2 415 & 2 438 &  46 800 &  47 700 &  47 250 \\
\rowcolor{gray!30} & ResNetIm & 0.188 & 0.369 &   -   &   340 &   347 &    -  & 338 400 & 332 100 &     -   \\ \hline 
                   &   ResNet & 0.069 & 0.115 & 0.205 &   928 & 1 113 & 1 249 &   4 993 &   4 161 &   3 709 \\ 
2 GPUs             &   2D-CNN & 0.018 & 0.029 & 0.051 & 3 556 & 4 414 & 5 020 &   1 303 &   1 049 &     923 \\ 
\rowcolor{gray!30} & CaffeNet & 0.037 & 0.045 & 0.071 & 1 730 & 2 844 & 3 606 &  66 600 &  40 500 &  31 950 \\
\rowcolor{gray!30} & ResNetIm & 0.099 & 0.181 & 0.337 &   646 &   707 &   760 & 178 200 & 162 900 & 151 650 \\ \hline 
                   &   ResNet & 0.053 & 0.105 & 0.184 &  1208 & 1 219 & 1 391 &   3 836 &   3 799 &   3 329 \\ 
4 GPUs             &   2D-CNN & 0.019 & 0.029 & 0.052 & 3 368 & 4 414 & 4 923 &   1 375 &   1 049 &     941 \\ 
\rowcolor{gray!30} & CaffeNet & 0.111 & 0.118 & 0.130 &   577 & 1 085 & 1 969 & 199 800 & 106 200 &  58 500 \\
\rowcolor{gray!30} & ResNetIm & 0.097 & 0.176 & 0.314 &   660 &   727 &   815 & 174 600 & 158 400 & 141 300 \\ \hline 
\end{tabular} 
\end{table}

\begin{table}   
\caption{Energy measurements (in joules) on a Pascal GPU for forward 
and backward steps of four CNN models using one, two and four GPUs 
and three batch sizes: 64, 128 and 256.}\label{t:pascal-energy}
\centering
\tabsize
\setlength{\tabcolsep}{0.5em}
\begin{tabular}{c|l|rrr|rrr|rrr}
\multicolumn{11}{c}{Forward} \\ \cline{3-11}
\multicolumn{2}{r|}{~} & \multicolumn{3}{c|}{Joules per batch} & \multicolumn{3}{c|}{Joules per second} & \multicolumn{3}{c}{Joules for whole training} \\ 
\multicolumn{2}{r|}{Batch size $\longrightarrow$} &  64 & 128 & 256 & 64 & 128 & 256 & 64 & 128 & 256 \\ \hline 
                   &   ResNet &  4.255 &  8.308 & 15.732 & 175 & 212 & 215 &    307 934 &    300 613 &    284 629 \\  
1 GPU              &   2D-CNN &  3.891 &  8.335 & 16.614 & 225 & 240 & 248 &    281 552 &    301 580 &    300 576 \\ 
\rowcolor{gray!30} & CaffeNet &  3.203 &  6.552 & 12.428 & 227 & 223 & 243 &  5 764 522 &  5 896 956 &  5 592 634 \\ 
\rowcolor{gray!30} & ResNetIm & 17.331 & 31.707 &    -   & 229 & 213 &  -  & 31 195 884 & 28 535 892 &      -     \\
\hline 
                   &   ResNet &  2.721 &  4.274 &  8.529 & 151 & 169 & 211 &    196 931 &    154 633 &    154 305 \\ 
2 GPUs             &   2D-CNN &  1.754 &  4.126 &  8.951 & 219 & 250 & 268 &    126 948 &    149 305 &    161 949 \\ 
\rowcolor{gray!30} & CaffeNet &  1.818 &  3.307 &  6.782 & 204 & 210 & 222 &  3 272 331 &  2 976 198 &  3 052 121 \\ 
\rowcolor{gray!30} & ResNetIm &  8.441 & 17.280 & 30.902 & 212 & 229 & 214 & 15 194 106 & 15 552 276 & 13 905 941 \\
\hline 
                   &   ResNet &  2.222 &  2.797 &  4.322 & 136 & 143 & 173 &    160 833 &    101 200 &     78 185 \\ 
4 GPUs             &   2D-CNN &  0.776 &  1.819 &  3.372 & 247 & 222 & 232 &     56 177 &     65 833 &     61 012 \\ 
\rowcolor{gray!30} & CaffeNet &  0.813 &  1.901 &  3.323 & 228 & 214 & 209 &  1 464 046 &  1 711 015 &  1 495 398 \\ 
\rowcolor{gray!30} & ResNetIm &  4.692 &  9.054 & 16.837 & 214 & 225 & 222 &  8 446 127 &  8 148 414 &  7 576 728 \\
\hline
\multicolumn{11}{c}{~} \\
\multicolumn{11}{c}{Backward} \\ \cline{3-11}
\multicolumn{2}{c|}{~} & \multicolumn{3}{c|}{Joules per batch} & \multicolumn{3}{c|}{Joules per second} & \multicolumn{3}{c}{Joules for whole training} \\ 
\multicolumn{2}{r|}{Batch size $\longrightarrow$} & 64 & 128 & 256 & 64 & 128 & 256 & 64 & 128 & 256 \\ \hline 
                   &   ResNet & 15.990 & 28.531 & 26.714 & 194 & 185 & 177 &  1 157 146 &  1 032 357 &    483 313 \\  
1 GPU              &   2D-CNN &  5.132 & 10.078 & 20.180 & 238 & 227 & 227 &    371 387 &    364 665 &    365 102 \\ 
\rowcolor{gray!30} & CaffeNet &  6.284 & 11.887 & 23.135 & 252 & 240 & 229 & 11 310 979 & 10 698 568 & 10 410 800 \\ 
\rowcolor{gray!30} & ResNetIm & 36.643 & 66.454 &   -    & 207 & 183 &  -  & 65 957 112 & 59 808 896 &     -    \\ \hline 
                   &   ResNet & 12.374 & 21.295 & 36.600 & 167 & 174 & 172 &    895 508 &    770 549 &    662 165 \\ 
2 GPUs             &   2D-CNN &  3.251 &  6.012 & 11.178 & 187 & 222 & 234 &    235 272 &    217 530 &    202 237 \\ 
\rowcolor{gray!30} & CaffeNet &  6.328 &  9.384 & 14.768 & 175 & 203 & 228 & 11 390 632 &  8 445 849 &  6 645 535 \\ 
\rowcolor{gray!30} & ResNetIm & 18.679 & 35.967 & 63.990 & 199 & 205 & 182 & 33 621 476 & 32 369 871 & 28 795 564 \\ \hline 
                   &   ResNet & 12.380 & 18.103 & 28.557 & 119 & 134 & 148 &    895 904 &    655 040 &    516 653 \\ 
4 GPUs             &   2D-CNN &  4.485 & 12.920 & 26.749 & 128 & 123 & 115 &    324 571 &    467 505 &    483 944 \\ 
\rowcolor{gray!30} & CaffeNet & 12.254 & 13.909 & 16.301 & 121 & 135 & 140 & 22 058 063 & 12 518 079 &  7 335 402 \\ 
\rowcolor{gray!30} & ResNetIm & 14.824 & 26.025 & 45.179 & 178 & 175 & 166 & 26 682 456 & 23 422 753 & 20 330 641 \\ \hline 
\end{tabular} 
\end{table}

\begin{table}   
\caption{Total execution time, energy consumption and EDP measurements 
for Pascal considering forward + backward steps of four CNN models using one, two and 
four GPUs and three batch sizes: 64, 128 and 256. Best results per measurement are 
boldfaced.}\label{t:edp}
\centering
\tabsize
\setlength{\tabcolsep}{0.6em}
\begin{tabular}{c|l|rrr|rrr|rrr}
\multicolumn{2}{c|}{~} & \multicolumn{3}{c|}{Kiloseconds (ks)} & \multicolumn{3}{c|}{Megajoules (MJ)} & \multicolumn{3}{c}{EDP} \\ 
\multicolumn{2}{r|}{Batch size $\longrightarrow$} &  64 & 128 & 256 & 64 & 128 & 256 & 64 & 128 & 256 \\ \hline 
                           & 1 GPU  &  11.1 &   9.4 &           5.8 &         1.47 &   1.33 &  \textbf{0.77}&    16.3 &    12.6 &     \textbf{4.5}\\  
ResNet                     & 2 GPUs &   6.8 &   5.2 &           4.5 &         2.18 &   1.85 &          1.63 &    14.9 &     9.7 &             7.3 \\ 
                           & 4 GPUs &   5.9 &   5.0 &   \textbf{4.1}&         4.30 &   3.05 &          2.28 &    25.5 &    15.1 &             9.2 \\
\hline 
                           & 1 GPU  &   2.5 &   2.8 &           2.7 & \textbf{0.65}&   0.67 &          0.67 &     1.6 &     1.9 &             1.8 \\  
2D-CNN                     & 2 GPUs &   1.9 &   1.6 &   \textbf{1.4}&         0.72 &   0.73 &          0.73 &     1.4 &     1.1 &     \textbf{1.0}\\ 
                           & 4 GPUs &   1.9 &   1.5 &   \textbf{1.4}&         1.62 &   1.81 &          1.68 &     3.1 &     2.7 &             2.3 \\
\hline 
\rowcolor{gray!30}         & 1 GPU  &  70.2 &  72.0 &          71.1 &        17.08 &  16.60 & \textbf{16.00}&  1198.7 &  1194.9 &          1137.8 \\  
\rowcolor{gray!30}CaffeNet & 2 GPUs &  81.0 &  53.1 &  \textbf{44.1}&        29.33 &  22.84 &         19.40 &  2375.4 &  1213.0 &   \textbf{855.3}\\ 
\rowcolor{gray!30}         & 4 GPUs & 212.4 & 118.8 &          69.3 &       101.26 &  60.97 &         38.39 & 21508.0 &  7243.5 &          2660.6 \\
\hline
\rowcolor{gray!30}         & 1 GPU  & 473.4 & 461.7 &       -       &        97.15 &  88.34 &       -       & 45992.2 & 40788.8 &         -       \\  
\rowcolor{gray!30}ResNetIm & 2 GPUs & 250.2 & 229.5 &         213.8 &        97.63 &  95.84 & \textbf{85.40}& 24427.3 & 21996.3 & \textbf{18254.9}\\ 
\rowcolor{gray!30}         & 4 GPUs & 237.6 & 214.2 & \textbf{191.7}&       146.39 & 132.87 &        120.61 & 34781.5 & 28460.5 &         23121.2 \\
\hline
\end{tabular} 
\end{table}

In this section we conduct experiments enabling the following hardware configurations: 
(1) one Pascal GPU , (2) two Pascal GPUs and (3) 2 Pascal + 2 Maxwell GPUs, 
with the measurement infrastructure always plugged to a Pascal GPU.

Table \ref{t:pascal} shows execution times with three batch sizes: 64, 128 and 256. 
For ResNetIm, the batch of 256 samples was not executed because it does not fit 
within the GPU memory. The whole training process corresponds to the number of epochs shown
in Table \ref{t:iterations}.

\subsubsection{Forward step}

According to the timing values (seconds per batch) shown in Table \ref{t:pascal} 
for the forward step on a  single GPU, the performance of 2D-CNN, CaffeNet and 
ResNetIm is very similar for different batch sizes. Best values for these networks 
are obtained for batch sizes of 64 (2D-CNN and CaffeNet) and 128 (ResNetIm). 
You realize much better when observing peak numbers in the column of samples 
processed per second. Also for one GPU, ResNet obtains larger performance gaps 
for different batch sizes, with poor results on small batches to reflect 
its dependency of aithmetic intensity. For the scalability of our implementations, 
using two GPUs 2D-CNN and Caffenet reach outstanding speedups (around 2.0x) 
on large batch sizes. Similar scores are attained by ResNetIm for any batch size, 
with its worst 1.6x speedup for the largest batch size. When moving to four GPUs, 
marginal improvements are seen, ranging from 2\% (ResNet for a 256 batch size) 
to 19\% (ResNetIm for 128 batch size).  This is because the time is determined 
by the slowest GPU on Maxwell devices.

Last columns in Table \ref{t:pascal} include the time required to compute all 
the forward iterations (shown in Table~\ref{t:iterations}) to perform a complete 
training. On a single GPU, smaller values are obtained for large batch sizes in 
ResNet and RestNetIm. However, a batch size of 64 is better for 2D-CNN and CaffeNet. 
Extending to twin GPUs, time is significantly reduced in all cases, and again, 
we find marginal gains on four GPUs, even with scenarios where execution times 
slowdown a bit.

Table \ref{t:pascal-energy} shows in three main columns the joules per batch, 
joules per second (watts) and joules for whole training spent in the forward 
and backward step for three batch sizes.
Note that power consumption is measured in one GPU. That way, when running the experiments 
on multiple GPUs, the energy spent must be multiplied by the number of GPUs to take 
into account all devices (that is on four GPUs, the total energy is the sum of two Pascals
and two Maxwells). Starting on a single GPU, joules per batch increase as the batch size 
grows because more samples per batch are processed. Likewise, the value of joules per second 
is higher in most cases for larger batch sizes as more computational density is available. 
Joules spent for the whole training not only depend on the joules per second value (watts), 
but also on the number of samples processed per second (that is, the execution time for the 
whole training process). When considering all these aspects for the forward step, ResNet, 
CaffeNet and ResNetIm networks give their best for the largest batch sizes, while 2D-CNN 
does it for a batch size of 128. On multi-GPUs, joules per second for a specific batch size 
is smaller when using more GPUs to reflect the distribution of samples per batch.

\subsubsection{Backward step}\label{s:backward-step}

The backward step takes longer than the forward step (see samples per second for each case). 
We identify a peculiar behaviour in ResNet for a batch size of $256$ on a single GPU, 
where seconds per batch are very similar to those of a batch size of $128$ (in principle, 
it should double those times). Using the Nvidia CUDA Profiler for a closer analysis, 
we found that the last 5 convolutions using a batch size of $128$ are executed with 
the function \textsf{wgrad\_alg0\_engine}, whereas for the batch size of $256$, 
those convolutions call the function \textsf{wgrad\_alg1\_engine}. 
Those functions are automatically included in the final code by cuDNN. 
Comparing their execution times, \textsf{wgrad\_alg1\_engine} is quite faster than 
\textsf{wgrad\_alg0\_engine} to benefit the batch size of $256$.
Using two GPUs, speedups are a bit lower than in the forward step, with the best value, 1.9x,
to be reached for ResNet, 2D-CNN and ResNetIm for batch sizes of 128, 256 and 128, respectively. 
That indicates a good overlapping between kernels computation and AllReduce transfers.  
In CaffeNet, which has the lowest CDC, the kernel computation cannot hide completely 
the data transfer performed by AllReduce and, consequently, the multi-GPU version for 
this network reduces its speed-up to 1.5x for a batch size of 256, and even worse, 
when using four GPUs. The only network that obtains a significant improvement using 
four devices is ResNet with up to a 30\% gain over the twin GPUs scenario.

Last columns in Table \ref{t:pascal} show the time spent to compute all backward iterations 
(as shown in Table\ref{t:iterations}) to perform a complete training. As already indicated 
in the forward step, on a single GPU, the process accelerates on lower batch sizes for ResNet 
and ResNetIm models. However, a batch size of 64 is better for 2D-CNN and CaffeNet. 
The use of two GPUs reduces the execution time in all forward and backward scenarios, 
but using four GPUs, only ResNet and ResNetIm improve during backward.

Table \ref{t:pascal-energy} shows the energy consumption for the backward step.
On a single GPU, ResNet and ResNetIm networks consume less energy on larger
batch sizes, whereas 2D-CNN and CaffeNet do it for a batch size of 64. 
For the multi-GPU case, joules per second for a specific batch size decrease 
when more GPUs are used, as the number of samples per batch is reduced and, 
consequently, less computation is performed.

\subsubsection{Setup comparison}

Now we compare all setups run on Pascal (i.e. number of GPUs, batch size and CNN models) 
to extract some conclusions regarding execution time, energy consumption and a combined 
metric of them, the Energy Delay Product (EDP)~\cite{laros2012book}.
Table~\ref{t:edp} summarizes those numbers for forward + backward steps during 
the complete training process. For a more compact representation, we measure time 
in kiloseconds (ks) and energy in Megajoules (MJ).

Focusing on execution time, the best option is a batch size of 256 samples 
for any network model. Depending on the type of architecture, it is better 
to use two GPUs for AlexNet-based models (2D-CNN and CaffeNet) and four GPUs 
for ResNet-based models (ResNet and ResNetIm).

For the energy exam, the best option is a batch size of 256 samples in most cases. 
Only with 2D-CNN it is better to use a small batch size of 64 samples, and only 
with ResNetIm we improve energy using two GPUs.

Finally, using the EDP metric, the best option overall is a large batch size 
with two GPUs. ResNet is an exception with one GPU as winner numbers due to the 
\textsf{wgrad\_alg1\_engine} problem already described in Section \ref{s:backward-step}.

\subsection{Results on Maxwell}\label{s:maxwell-results}  

\begin{table}   
\caption{Execution times (in seconds) on a Maxwell GPU for forward and backward steps 
of four CNN models using one, two and four GPUs and three batch sizes: 64, 
128 and 256.}\label{t:maxwell}
\centering
\tabsize
\begin{tabular}{c|l|rrr|rrr|rrr}
\multicolumn{11}{c}{Forward} \\ \cline{3-11}
\multicolumn{2}{c|}{~} & \multicolumn{3}{c|}{Seconds per batch} & \multicolumn{3}{c|}{Samples per second} & \multicolumn{3}{c}{Seconds for whole training} \\ 
\multicolumn{2}{r|}{Batch size $\longrightarrow$} &  64 & 128 & 256 & 64 & 128 & 256 & 64 & 128 & 256 \\ \hline 
                   &   ResNet & 0.039 & 0.059 & 0.094 & 1 641 &  2 169 &  2 723 &   2 822 &   2 135 &  1 701 \\  
1 GPU              &   2D-CNN & 0.026 & 0.049 & 0.097 & 2 462 &  2 612 &  2 639 &   1 882 &   1 773 &  1 755 \\ 
\rowcolor{gray!30} & CaffeNet & 0.023 & 0.045 & 0.090 & 2 783 &  2 844 &  2 844 &  41 400 &  40 500 & 40 500 \\
\rowcolor{gray!30} & ResNetIm & 0.110 & 0.211 &   -   &   582 &    607 &    -   & 198 000 & 189 900 &    -   \\ \hline 
                   &   ResNet & 0.031 & 0.039 & 0.059 & 2 065 &  3 282 &  4 339 &   2 243 &   1 411 &  1 067 \\ 
2 GPUs             &   2D-CNN & 0.014 & 0.025 & 0.048 & 4 571 &  5 120 &  5 333 &   1 013 &     905 &    868 \\ 
\rowcolor{gray!30} & CaffeNet & 0.014 & 0.024 & 0.046 & 4 571 &  5 333 &  5 565 &  25 200 &  21 600 & 20 700 \\
\rowcolor{gray!30} & ResNetIm & 0.059 & 0.107 & 0.209 & 1 085 &  1 196 &  1 225 & 106 200 &  96 300 & 94 050 \\ \hline 
                   &   ResNet & 0.029 & 0.032 & 0.040 & 2 207 &  4 000 &  6 400 &    2099 &   1 158 &    724 \\ 
4 GPUs             &   2D-CNN & 0.007 & 0.012 & 0.023 & 9 143 & 10 667 & 11 130 &     507 &     434 &    416 \\ 
\rowcolor{gray!30} & CaffeNet & 0.007 & 0.014 & 0.024 & 9 143 &  9 143 & 10 667 &  12 600 &  12 600 & 10 800 \\
\rowcolor{gray!30} & ResNetIm & 0.035 & 0.062 & 0.112 & 1 829 &  2 065 &  2 286 &  63 000 &  5 5800 & 50 400 \\ \hline 
\multicolumn{11}{c}{~} \\
\multicolumn{11}{c}{Backward} \\ \cline{3-11}
\multicolumn{2}{c|}{~} & \multicolumn{3}{c|}{Seconds per batch} & \multicolumn{3}{c|}{Samples per second} & \multicolumn{3}{c}{Seconds for whole training} \\ 
\multicolumn{2}{r|}{Batch size $\longrightarrow$} &  64 & 128 & 256 & 64 & 128 & 256 & 64 & 128 & 256 \\ \hline 
                   &   ResNet & 0.110 & 0.184 & 0.336 &   582 &   696 &   762 &   7 960 &   6 658 &   6 079 \\  
1 GPU              &   2D-CNN & 0.039 & 0.074 & 0.148 & 1 641 & 1 730 & 1 730 &   2 822 &   2 678 &   2 678 \\ 
\rowcolor{gray!30} & CaffeNet & 0.047 & 0.092 & 0.182 & 1 362 & 1 391 & 1 407 &  84 600 &  82 800 &  81 900 \\
\rowcolor{gray!30} & ResNetIm & 0.269 & 0.520 &   -   &   238 &   246 &   -   & 484 200 & 468 000 &     -   \\ \hline 
                   &   ResNet & 0.049 & 0.078 & 0.135 & 1 306 & 1 641 & 1 896 &   3 546 &   2 822 &   2 442 \\ 
2 GPUs             &   2D-CNN & 0.025 & 0.045 & 0.080 & 2 560 & 2 844 & 3 200 &   1 809 &   1 628 &   1 447 \\ 
\rowcolor{gray!30} & CaffeNet & 0.044 & 0.060 & 0.103 & 1 455 & 2 133 & 2 485 &  79 200 &  54 000 &  46 350 \\
\rowcolor{gray!30} & ResNetIm & 0.141 & 0.261 & 0.512 &   454 &   490 &   500 & 253 800 & 234 900 & 230 400 \\ \hline 
                   &   ResNet & 0.053 & 0.105 & 0.184 & 1 208 & 1 219 & 1 391 &   3 836 &   3 799 &   3 329 \\ 
4 GPUs             &   2D-CNN & 0.019 & 0.029 & 0.052 & 3 368 & 4 414 & 4 923 &   1 375 &   1 049 &     941 \\ 
\rowcolor{gray!30} & CaffeNet & 0.111 & 0.118 & 0.130 &   577 & 1 085 & 1 969 & 199 800 & 106 200 &  58 500 \\
\rowcolor{gray!30} & ResNetIm & 0.097 & 0.176 & 0.314 &   660 &   727 &   815 & 174 600 & 158 400 & 141 300 \\ \hline 
\end{tabular} 
\end{table}

\begin{table}   
\caption{Energy measurements (in joules) on a Maxwell GPU for forward 
and backward steps of four CNN models using one, two and four GPUs 
and three batch sizes: 64, 128 and 256.}\label{t:maxwell-energy}
\centering
\tabsize
\setlength{\tabcolsep}{0.5em}
\begin{tabular}{c|l|rrr|rrr|rrr}
\multicolumn{11}{c}{Forward} \\ \cline{3-11}
\multicolumn{2}{c|}{~} & \multicolumn{3}{c|}{Joules per batch} & \multicolumn{3}{c|}{Joules per second} & \multicolumn{3}{c}{Joules for whole training} \\ 
\multicolumn{2}{r|}{Batch size $\longrightarrow$} &  64 & 128 & 256 & 64 & 128 & 256 & 64 & 128 & 256 \\ \hline 
                   &   ResNet &  6.111 & 11.070 & 19.342 & 163 & 174 & 176 &    442 221 &    400 540 &    349 944 \\  
1 GPU              &   2D-CNN &  6.248 & 12.199 & 24.432 & 208 & 208 & 214 &    452 148 &    441 402 &    442 022 \\ 
\rowcolor{gray!30} & CaffeNet &  5.379 & 10.499 & 21.217 & 209 & 206 & 215 &  9 681 859 &  9 448 940 &  9 547 708 \\ 
\rowcolor{gray!30} & ResNetIm & 22.864 & 45.341 &    -   & 201 & 203 &  -  & 41 155 296 & 40 807 243 &      -     \\ \hline 
                   &   ResNet &  3.988 &  6.188 & 11.183 & 154 & 156 & 172 &    288 620 &    223 899 &    202 325 \\ 
2 GPUs             &   2D-CNN &  3.052 &  6.272 & 12.221 & 194 & 213 & 217 &    220 855 &    226 959 &    221 108 \\ 
\rowcolor{gray!30} & CaffeNet &  2.995 &  5.255 & 10.615 & 185 & 196 & 203 &  5 391 093 &  4 729 650 &  4 776 933 \\ 
\rowcolor{gray!30} & ResNetIm & 12.581 & 22.611 & 45.459 & 226 & 199 & 206 & 22 646 534 & 20 349 928 & 20 456 374 \\ \hline 
                   &   ResNet &  2.979 &  4.075 &  6.350 & 141 & 152 & 156 &    215 574 &    147 466 &    114 892 \\ 
4 GPUs             &   2D-CNN &  1.530 &  3.146 &  6.003 & 192 & 207 & 219 &    110 705 &    113 847 &    108 603 \\ 
\rowcolor{gray!30} & CaffeNet &  1.493 &  3.103 &  5.381 & 193 & 206 & 204 &  2 686 577 &  2 792 498 &  2 421 440 \\ 
\rowcolor{gray!30} & ResNetIm &  6.043 & 11.620 & 21.899 & 190 & 188 & 192 & 10 876 855 & 10 457 805 &  9 854 684 \\ \hline
\multicolumn{11}{c}{~} \\
\multicolumn{11}{c}{Backward} \\ \cline{3-11}
\multicolumn{2}{c|}{~} & \multicolumn{3}{c|}{Joules per batch} & \multicolumn{3}{c|}{Joules per second} & \multicolumn{3}{c}{Joules for whole training} \\ 
\multicolumn{2}{r|}{Batch size $\longrightarrow$} &  64 & 128 & 256 & 64 & 128 & 256 & 64 & 128 & 256 \\ \hline 
                   &   ResNet & 13.012 &  23.047 &  45.465 & 166 & 172 & 178 &    941 655 &    833 920 &    822 555 \\  
1 GPU              &   2D-CNN &  8.245 &  16.542 &  33.743 & 210 & 214 & 223 &    596 691 &    598 569 &    610 472 \\ 
\rowcolor{gray!30} & CaffeNet &  9.995 &  20.007 &  40.783 & 217 & 210 & 218 & 17 990 236 & 18 006 323 & 18 352 400 \\ 
\rowcolor{gray!30} & ResNetIm & 51.547 & 101.679 &     -   & 191 & 190 &  -  & 92 783 786 & 91 511 546 &     -      \\ \hline 
                   &   ResNet &  8.616 &  14.233 &  24.921 & 156 & 157 & 165 &    623 520 &    515 010 &    450 874 \\ 
2 GPUs             &   2D-CNN &  4.634 &   8.714 &  17.224 & 180 & 202 & 214 &    335 346 &    315 321 &    311 623 \\ 
\rowcolor{gray!30} & CaffeNet &  6.935 &  11.803 &  21.529 & 170 & 187 & 208 & 12 483 290 & 10 622 845 &  9 688 155 \\ 
\rowcolor{gray!30} & ResNetIm & 29.857 &  50.594 & 100.843 & 208 & 189 & 195 & 53 742 006 & 45 534 792 & 45 379 223 \\ \hline 
                   &   ResNet & 12.123 &  17.221 &  23.668 & 123 & 124 & 127 &    877 320 &    623 140 &    428 193 \\ 
4 GPUs             &   2D-CNN &  4.424 &   7.095 &  10.440 & 137 & 148 & 180 &    320 158 &    256 728 &    188 872 \\ 
\rowcolor{gray!30} & CaffeNet & 13.568 &  14.960 &  17.653 & 134 & 142 & 154 & 24 422 178 & 13 464 314 &  7 943 784 \\ 
\rowcolor{gray!30} & ResNetIm & 15.104 &  27.117 &  50.097 & 177 & 179 & 182 & 27 187 992 & 24 405 527 & 22 543 695 \\ \hline 
\end{tabular} 
\end{table}

\begin{table}   
\caption{Execution time, energy consumption and EDP measurements for 
Maxwell architecture considering forward + backward steps of four CNN models using one, 
two and four GPUs and three batch sizes: 64, 128 and 256. Best results per measurement 
are marked in bold.}\label{t:edp-maxwell}
\centering
\tabsize
\setlength{\tabcolsep}{0.6em}
\begin{tabular}{c|l|rrr|rrr|rrr}
\multicolumn{2}{c|}{~} & \multicolumn{3}{c|}{Kiloseconds (ks)} & \multicolumn{3}{c|}{Megajoules (MJ)} & \multicolumn{3}{c}{EDP} \\ 
\multicolumn{2}{r|}{Batch size $\longrightarrow$} &  64 & 128 & 256 & 64 & 128 & 256 & 64 & 128 & 256 \\ \hline 
                           & 1 GPU  &  10.8 &   8.8 &           7.8 &   1.38 &          1.23 &   \textbf{1.17}&    14.9 &    10.9 &             9.1 \\  
ResNet                     & 2 GPUs &   5.8 &   4.2 &   \textbf{3.5}&   1.82 &          1.48 &           1.31 &    10.6 &     6.3 &     \textbf{4.6}\\ 
                           & 4 GPUs &   5.9 &   5.0 &           4.1 &   4.30 &          3.05 &           2.28 &    25.5 &    15.1 &             9.2 \\ \hline 
                           & 1 GPU  &   4.7 &   4.5 &           4.4 &   1.05 &  \textbf{1.04}&           1.05 &     4.9 &     4.6 &             4.7 \\  
2D-CNN                     & 2 GPUs &   2.8 &   2.5 &           2.3 &   1.11 &          1.08 &           1.07 &     3.1 &     2.7 &             2.5 \\ 
                           & 4 GPUs &   1.9 &   1.5 &   \textbf{1.4}&   1.62 &          1.81 &           1.68 &     3.1 &     2.7 &     \textbf{2.3}\\ \hline 
\rowcolor{gray!30}         & 1 GPU  & 126.0 & 123.3 &         122.4 &  27.67 & \textbf{27.46}&          27.90 &  3486.7 &  3385.2 &          3415.0 \\  
\rowcolor{gray!30}CaffeNet & 2 GPUs & 104.4 &  75.6 &  \textbf{67.1}&  35.75 &         30.70 &          28.93 &  3732.2 &  2321.3 &  \textbf{1939.8}\\ 
\rowcolor{gray!30}         & 4 GPUs & 212.4 & 118.8 &          69.3 & 101.26 &         60.97 &          38.39 & 21508.0 &  7243.5 &          2660.6 \\ \hline
\rowcolor{gray!30}         & 1 GPU  & 682.2 & 657.9 &        -      & 133.94 &        132.32 &       -        & 91373.2 & 87052.5 &        -        \\  
\rowcolor{gray!30}ResNetIm & 2 GPUs & 360.0 & 331.2 &         324.5 & 152.78 &        131.77 &         131.67 & 54999.7 & 43642.0 &         42720.7 \\ 
\rowcolor{gray!30}         & 4 GPUs & 237.6 & 214.2 & \textbf{191.7}& 146.39 &        132.87 & \textbf{120.61}& 34781.5 & 28460.5 & \textbf{23121.2}\\ \hline
\end{tabular} 
\end{table}

In this section, experiments are conducted using the following hardware configurations: 
(1) one Maxwell GPU, (2) two Maxwell GPUs and (3) 2 Pascal + 2 Maxwell GPUs, 
with the measurement infrastructure always plugged to a Maxwell GPU.

Table \ref{t:maxwell} shows execution times for three batch sizes: 64, 128 and 256. 
Again, the batch of 256 samples has not been executed for ResNetIm because it exceeds
the GPU global memory size.

\subsubsection{Forward step}

Execution times follow a similar pattern to that already analyzed for Pascal, 
but as expected there is a general slowdown in performance because Maxwell 
is an older device. In the forward step, 2D-CNN,  CaffeNet and ResNetIm all exhibit 
a stable throughput (samples per second) for any batch size, 
whereas ResNet reaches its peak for a batch size of 256.
For two GPUs, excellent speedups values close to the optimal 2x are obtained for 2D-CNN, 
CaffeNet and ResNetImIn. A lower gain is achieved by ResNet. The configuration 
with four GPUs also produces excellent results with speeds around 4x for 2D-CNN and CaffeNet.

Table \ref{t:maxwell-energy} shows power results measured following the same methodology
described for Pascal. Again, the energy budget correlates with the execution time, but now
power savings are smaller, basically because Maxwell was manufactured on 28 nm. transistors
(Pascal benefits from a 16 nm. technology). For the forward step, lower energy requirements
are typically achieved for larger batch sizes.

\subsubsection{Backward step}

Here, values on a single GPU follow tendencies already shown for the forward case. 
Twin GPUs exhibit good performance numbers for specific batch sizes, specially with ResNet 
and ResNetIm, to indicate effective overlapping between kernels computation and Allreduce 
transfers. Speedup values on four GPUs are modest, with improvements just for 2D-CNN and ResNetIm,
and for the energy consumption, optimal values are found on larger batch sizes.

Table \ref{t:maxwell-energy} shows power results measured following the same methodology
described for Pascal. During the backward step, the benefit of 
using multiple GPUs during training is more clear compared to Pascal. Finally, comparing 
the whole process, forward plus backward, energy requirements are optimized on larger batches
for all networks, what correlates with performance.

\subsubsection{Setup comparison}

As for Pascal, we compare here all setups run in Maxwell to draw some conclusions.
Table~\ref{t:edp-maxwell} compiles all required numbers.

Starting with the execution time, again the best option is a batch size of 256 samples. 
Depending on the type of architecture, it is better to use two GPUs with network models 
having lower CTC (i.e. ResNet and CaffeNet) and 4 GPUs for the networks with higher CTC 
(i.e. 2D-CNN and ResNetIm).

For the energy discussion, the best option is a batch size of 256 samples with ResNet-based 
models (ResNet/ResNetIm) and 128 samples with AlexNet-based models (2D-CNN/CaffeNet). 
And best records are registered on a single GPU, with the exception of ResNetIm on 4 GPUs.

Finally, for the EDP metric, the best option is a large batch size with two GPUs for networks 
with low CTC. For higher CTC values, optimal values are found on larger batch size 
using four GPUs.

\subsection{Pascal versus Maxwell}\label{s:PascalvsMaxwell}   

\begin{figure}   
\centering
\includegraphics[width=\textwidth]{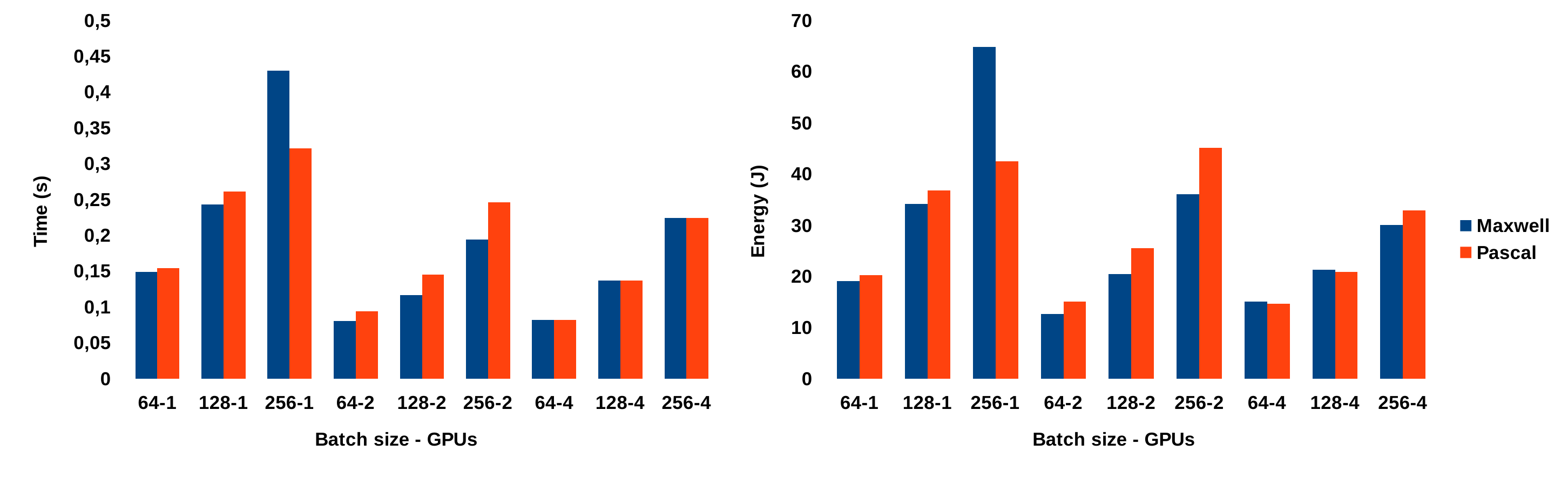}
\caption{Comparative between Maxwell and Pascal for ResNet. 
(left) time per batch (right) joules per batch. 
All measurements are for forward + backward steps using 1, 2 and 4 GPUs 
and three batch sizes: 64, 128 and 256.}
\label{f:comp-resnet}
\end{figure}

\begin{figure}   
\centering
\includegraphics[width=\textwidth]{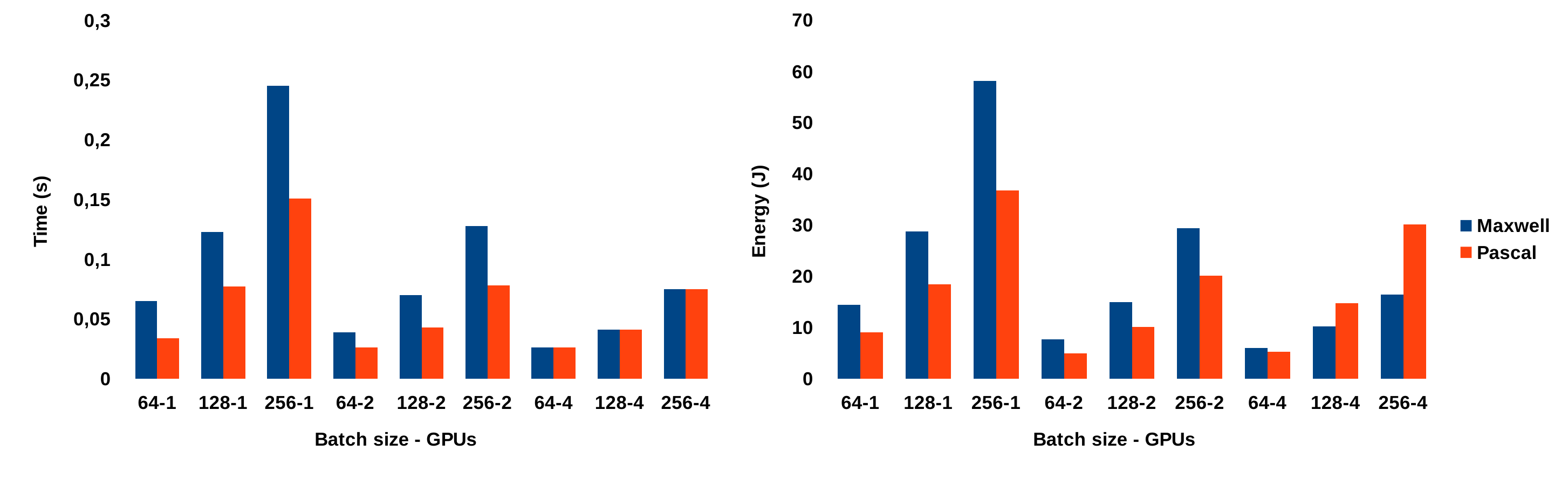}
\caption{Comparative between Maxwell and Pascal for 2D-CNN. 
Left: time per batch. Right: joules per batch. 
All measurements are for forward + backward steps using 1, 2 and 4 GPUs
and three batch sizes: 64, 128 and 256.}
\label{f:comp-2d}
\end{figure}

\begin{figure}   
\centering
\includegraphics[width=\textwidth]{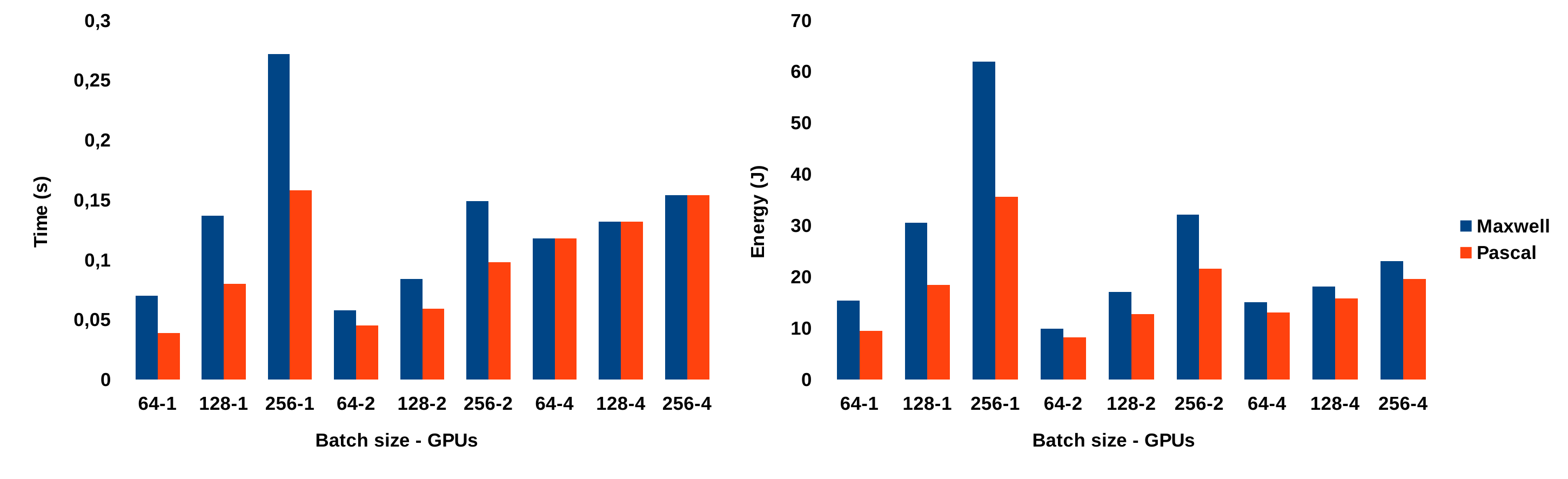}
\caption{Comparative between Maxwell and Pascal for CaffeNet. 
Left: time per batch. Right: joules per batch. 
All measurements are for forward + backward steps using 1, 2 and 4 GPUs 
and three batch sizes: 64, 128 and 256.}
\label{f:comp-caffenet}
\end{figure}

\begin{figure}   
\centering
\includegraphics[width=\textwidth]{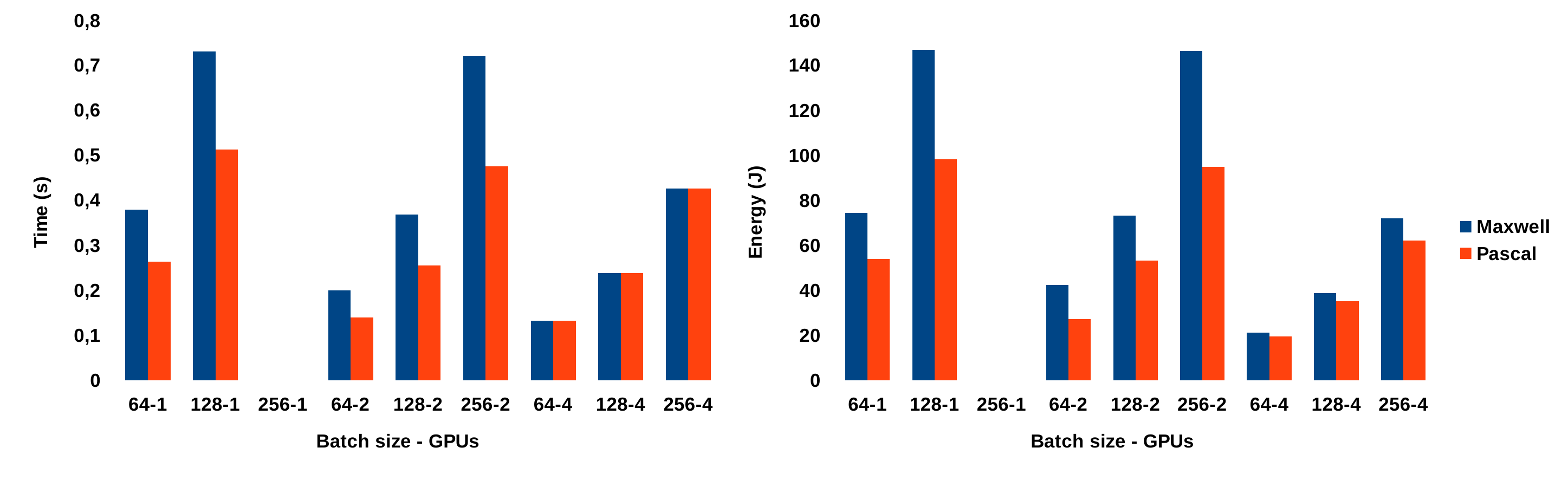}
\caption{Comparative between Maxwell and Pascal for ResNetIm. 
Left: time per batch. Right: joules per batch. 
All measurements are for forward + backward steps using 1, 2 and 4 GPUs 
and three batch sizes: 64, 128 and 256.}
\label{f:comp-resnetim}
\end{figure}

We now want to compare the execution time and power consumption of our four CNN models 
for all batch sizes and number of GPUs on Maxwell and Pascal. These results are 
compiled in Figures \ref{f:comp-resnet},~\ref{f:comp-2d},~\ref{f:comp-caffenet} 
and~\ref{f:comp-resnetim}. For the bar names in our charts, we follow the rule 
\textit{batch\_size-number\_GPUs}. For example, 64-4 stands for a batch of 64 elements 
using 4 GPUs. Again, there are fluctuations in 4 GPUs because the two Pascals 
have to wait the two Maxwells to conclude. This effect can only be seen in 
energy consumption because the time included in the plots is the slowest of all GPUs 
(that is, time measurements for Maxwell match those of Pascal when using 4 GPUs).

Results indicate that for 2D-CNN, CaffeNet and ResNetIm, Pascal is ahead in
performance and power consumption. The improvement can be quantified within a $30-40\%$ 
range depending on the experiment. For ResNet, performance drops in Pascal versus Maxwell 
when using multiple GPUs, and also on a single GPU for small batch sizes.
To explain this behaviour, we have added Table \ref{t:comp-resnet} specifically 
for ResNet, covering all batch sizes and GPUs. As we can see, during forward, 
the behaviour is around $30\%$ better in Pascal. However, during backward, 
its performance decreases heavily and Maxwell overtakes it. We have found Pascal 
to be affected by the algorithm change used within cuDNN to compute the last 
convolutions of this model 
(\textsf{wgrad\_alg0\_engine} method instead of \textsf{wgrad\_alg1\_engine}, 
being the former 25 times slower than the latter - see Section \ref{s:Pascal-results}). 
The only value which is not affected by this anomaly in Table \ref{t:comp-resnet} 
is the batch size of $256$ executed in one GPU (when using more GPUs, the batch size 
is distributed among them, and the threshold for the algorithm to switch to 
the swift version is never reached). This way, results worsen when using Pascal 
with ResNet if the threshold in the batch size is not reached, 
regardless of the number of GPUs. We are confident this anomaly will be solved 
in future releases of cuDNN to end up with faster executions like all those we have 
introduced here using the \textsf{wgrad\_alg1\_engine} method.

\begin{table}  
\caption{A comparison between Maxwell and Pascal during forward 
and backward steps using ResNet.}\label{t:comp-resnet}
\centering
\tabsize
\begin{tabular}{c|l|rrr|rrr}
\multicolumn{8}{c}{Forward} \\ \cline{2-8} 
       & & \multicolumn{3}{c|}{Time per batch} & \multicolumn{3}{c}{Joules per batch} \\
       & Batch size & Maxwell & Pascal & $\%$ & Maxwell & Pascal & $\%$ \\ \hline 
       &         64 &   0.039 &  0.029 & 25.6 &   6.111 &  4.255 & 30.4 \\  
1 GPU  &        128 &   0.059 &  0.041 & 30.5 &  11.070 &  8.308 & 24.9 \\ 
       &       256  &   0.094 &  0.067 & 28.7 &  19.342 & 15.732 & 18.7 \\ \hline 
       &         64 &   0.031 &  0.025 & 19.4 &   3.988 &  2.721 & 31.8 \\ 
2 GPUs &        128 &   0.039 &  0.030 & 23.1 &   6.188 &  4.274 & 30.9 \\ 
       &        256 &   0.059 &  0.041 & 30.5 &  11.183 &  8.529 & 23.7 \\ \hline 
       &         64 &   0.029 &  0.029 &    0 &   2.979 &  2.222 & 25.4 \\ 
4 GPUs &        128 &   0.032 &  0.032 &    0 &   4.075 &  2.797 & 31.4 \\ 
       &        256 &   0.040 &  0.040 &    0 &   6.350 &  4.322 & 31.9 \\ \hline 
\multicolumn{8}{c}{~} \\
\multicolumn{8}{c}{Backward} \\ \cline{2-8}
       & & \multicolumn{3}{c|}{Time per batch} & \multicolumn{3}{c}{Joules per batch} \\ 
       & Batch size & Maxwell & Pascal & $\%$ & Maxwell & Pascal & $\%$ \\ \hline 
       &        64  & 0.110 & 0.125 & -13.6 & 13.012 & 15.990 & -22.9 \\ 
1 GPU  &       128  & 0.184 & 0.220 & -19.6 & 23.047 & 28.531 & -23.8 \\ 
       &       256  & 0.336 & 0.254 &  24.4 & 45.465 & 26.714 &  41.2 \\ \hline 
       &        64  & 0.049 & 0.069 & -40.8 &  8.616 & 12.374 & -43.6 \\ 
2 GPUs &       128  & 0.078 & 0.115 & -47.4 & 14.233 & 21.295 & -49.6 \\ 
       &       256  & 0.135 & 0.205 & -51.9 & 24.921 & 36.600 & -46.9 \\ \hline 
       &        64  & 0.053 & 0.053 &     0 & 12.123 & 12.380 &  -2.1 \\ 
4 GPUs &       128  & 0.105 & 0.105 &     0 & 17.221 & 18.103 &  -5.1 \\ 
       &       256  & 0.184 & 0.184 &     0 & 23.668 & 28.557 & -20.7 \\ \hline 
\end{tabular} 
\end{table}

\subsection{Energy versus performance}\label{s:EnergyvsPerf}    

In general, the GPU evolution has demonstrated that performance does not correlate 
ideally with energy efficiency, because sometimes you experience severe 
power penalties when being eager on performance. In fact, Nvidia introduced 
GPU Boost and clock monitoring in Kepler GPUs back in 2012 to keep an eye on power 
at run time depending on computational requirements driven by every particular 
application. Later in 2014, when they released Maxwell, it was announced as 
the most power efficient GPU ever built \cite{GTX980}. Compared to its predecessor 
Kepler, multiprocessors were reduced to 128 cores and layout was reorganized 
into quadrants to shorten wires length. Communications and power lines were 
identified primary factors in energy consumption, so it was no surprise to find 
Maxwell ahead a 2x factor in performance per watt.
 
Enhancements introduced in 2016 with Pascal were driven by performance and energy, 
but with certain tradeoffs versus Maxwell. Focusing on Titan models to be fair, 
Table \ref{t:HW} summarizes features for the two GPUs used in our study. 
The Maxwell model contains 3072 cores at 1392 MHz clock rate, whereas the Pascal 
counterpart has 3584 cores running at 1911 MHz. The number of transistors on a chip 
and its frequency affect power in a linear way, which leads us to estimate Pascal 
around 65\% higher on energy demand, and presumably a similar percentage ahead 
in performance. When you increase wattage but reduce seconds proportionally, 
the energy toll in joules should remain constant, but there were good news for Pascal 
on a performance-per-watt basis: 
Multiprocessors were reduced to 64 cores and, overall, manufacturing process 
evolved from planar 28 nm. transistors to 16 nm. fin-FET ones \cite{P100}.
With those many variables affecting power and all side-effects among them, 
it is complex to assess pros and cons to determine a winner of the energy battle, 
and even more challenging to put differences in raw numbers.

Our set of experiments may shed some light driven by praxis. Table \ref{t:gflops} 
illustrates performance per watt on a wide number of settings, changing CNN models 
and batch sizes. Peak numbers are reached on one GPU computing the 2D-CNN model, 
where numbers are stable around 11 GFLOPS/w for Pascal and 7 GFLOPS/w for Maxwell 
(around 60\% deficit). 2D-CNN is also the optimal model on two GPUs, keeping 
distances between twin Pascals and twin Maxwells around 50\%. 
Official peak differences published by Nvidia in double precision numbers are 40\% 
(20 GFLOPS/w for an average Pascal GPU and 12 GFLOPS/w for the Maxwell counterpart), 
what tells us that we have found CNN models where those differences widen up to 
an additional 20\% regardless of the batch size chosen. We also see that energy 
efficiency is very sensitive to the CNN model computed, because there are other cases, 
like the ResNet model, where differences shorten very much among GPUs.

Our numbers also validate Nvidia estimations, because our global average 
for all models and batch sizes is 42\% running on a single GPU and 35\% when 
using a pair. The 4 GPUs setup may look confusing at first sight, but note
that we are not comparing 4 Pascals versus 4 Maxwells. Instead, we always use 
2 Pascals plus 2 Maxwells, that is, it is always the same run, just changing 
power measurements from one generation to another. That way, synchronizations 
may relax the faster twin Pascals to end up with similar power requirements 
versus the twins Maxwells. In other words, performance is mainly responsible 
for energy savings when running CNNs on Pascal, and the set of experiments 
gathered in this paper encourage you to press the throttle because 
you will not end up paying more on fuel.

In addition, we can see that CNN applications stay, in general, far from optimal 
performance per watt ratios: The maximum values we were able to attain are 11 GFLOPS/w 
on a Pascal and 7 GFLOPS/w on a Maxwell, whereas SGEMM (Single Precision General 
Matrix Multiply) reaches 42 GFLOPS/w in Pascal and 23 GFLOPS/w in Maxwell.
That means that we barely squeeze 25\% of the performance efficiency exhibited by 
a typical compute bound procedure. We expect this margin to shrink when using 
the new half data types that Nvidia introduced in Pascal particularly to benefit 
deep learning applications.

Finally, if we focus our analysis on the influence of the batch size, optimal 
performance and minimum energy consumption due to savings in training time 
are attained when increasing batch sizes as much as possible in all GPU scenarios. 
But there are a number of concerns regarding accuracy and datasets which deserve 
a closer attention. We address those two in sections \ref{s:accuracy} and 
\ref{s:best-approach}, respectively.

\begin{table}  
\caption{GFLOPS per watt for our four CNN models with three batch sizes 
measured on Pascal and Maxwell using 1, 2 and 4 GPUs (note that the 4 GPUs 
run is the same for Pascal and Maxwell, we just change the device where energy 
is measured). Results including both forward and backward. Averages are calculated 
per row and column, and last column reflects differences based on those.}\label{t:gflops}
\centering
\tabsize
\begin{tabular}{c|l|rrr|r|ccc|c|r}
\hline
\multicolumn{2}{c|}{GFLOPS/w measured on $\longrightarrow$} & \multicolumn{4}{c|}{Pascal} & \multicolumn{4}{c|}{Maxwell} & Pascal \\
\cline{3-10} 
\multicolumn{2}{r|}{Batch size $\longrightarrow$} & 64 & 128 & 256 & Average & 64 & 128 & 256 & Average & gain \\ 
\hline 
                                        &      ResNet   &      2.7 &      3.0 &      5.1 &      4.3 &      2.8 &      3.2 &      3.4 &      3.1 &       38\% \\ 
1 GPU                                   &      2D-CNN   &     11.1 &     10.9 &     10.9 &     11.0 &      6.9 &      7.0 &      6.9 &      6.9 &       59\% \\ 
\rowcolor{gray!30} (1 Pascal or         &      CaffeNet &      9.8 &     10.1 &     10.5 &     10.1 &      6.1 &      6.1 &      6.0 &      6.1 &       65\% \\ 
\rowcolor{gray!30}  1 Maxwell)          &      ResNetIm &      4.9 &      5.4 &     -    &      5.1 &      3.6 &      3.6 &     -    &      3.6 &       41\% \\ \cline{2-11}
                                        & {\bf Average} & {\bf 7.1}& {\bf 7.3}& {\bf 8.8}& {\bf 7.7}& {\bf 5.8}& {\bf 5.0}& {\bf 5.4}& {\bf 5.4}& {\bf +42\%}\\ \hline
\multicolumn{9}{c}{\vspace*{1.5mm}} \\ \hline
                                        &      ResNet   &      1.8 &      2.1 &      2.4 &      2.1 &      2.2 &      2.7 &      3.0 &      2.6 &      -20\% \\
2 GPUs                                  &      2D-CNN   &     10.0 &      9.9 &     10.0 &     10.0 &      6.5 &      6.7 &      6.8 &      6.7 &       49\% \\
\rowcolor{gray!30} (2 Pascals or        &      CaffeNet &      5.7 &      7.3 &      8.6 &      7.2 &      4.7 &      5.5 &      5.8 &      5.3 &       35\% \\
\rowcolor{gray!30}  2 Maxwells)         &      ResNetIm &      4.9 &      5.0 &      5.6 &      5.2 &      3.1 &      3.6 &      3.6 &      3.4 &       52\% \\ \cline{2-11}
                                        & {\bf Average} & {\bf 5.6}& {\bf 6.1}& {\bf 6.6}& {\bf 6.1}& {\bf 4.1}& {\bf 4.6}& {\bf 4.8}& {\bf 4.5}& {\bf +35\%}\\ \hline
\multicolumn{9}{c}{\vspace*{1.5mm}} \\ \hline
                                        &      ResNet   &      0.9 &      1.3 &      1.7 &      1.3 &      0.9 &      1.3 &      1.8 &      1.3 &        0\% \\
4 GPUs                                  &      2D-CNN   &      4.8 &      3.4 &      3.3 &      4.2 &      4.2 &      4.9 &      6.1 &      5.0 &      -16\% \\
\rowcolor{gray!30} (2 Pascals {\bf and} &      CaffeNet &      1.8 &      2.9 &      4.7 &      3.1 &      1.5 &      2.6 &      4.0 &      2.7 &       14\% \\
\rowcolor{gray!30} 2 Maxwells)          &      ResNetIm &      3.4 &      3.8 &      4.3 &      3.8 &      3.1 &      3.4 &      3.7 &      3.4 &       11\% \\ \cline{2-11}
                                        & {\bf Average} & {\bf 2.8}& {\bf 2.9}& {\bf 3.5}& {\bf 3.1}& {\bf 2.4}& {\bf 3.0}& {\bf 3.9}& {\bf 3.1}&        0\% \\ \hline
\end{tabular} 
\end{table}

\subsection{Accuracy}\label{s:accuracy}   

We extend our CNN analysis from performance and energy viewpoints in this section
to find a good model in terms of accuracy. In our experiments, we only consider
the batch size as tunable hyper-parameter, because all remaining ones have been taken 
from previously trained models with good accuracy. According to Caffe's implementation, 
the training with one or more GPUs leads to the same results, so we do not move
the number of GPUs. Moreover, we distinguish results taken from models using videos 
as input (Table \ref{t:acc}) from those using images (Table \ref{t:acc-imagenet}).

Table \ref{t:acc} summarizes the accuracy results for ResNet and 2D-CNN, where we can see
that the best model is 2D-CNN with $86.0\%$ of accuracy. On the other hand, the best ResNet 
model obtains a disappointing $76\%$. Overfitting is responsible for this low accuracy. 
This model contains a vast number of parameters, while the amount of training data available 
in the dataset is relatively small. Therefore, the model is not able to generalize to the test data. 
Comparing the accuracy among batch sizes in both models, the precision decreases with bigger 
batches because the average gradients are less noisy and the exploration capacity of the algorithm 
is reduced. On the other hand, with small batches, the algorithm explores better the solution space 
and, consequently, finds a better local minimum. This effect is more clear in ResNet due to the 
huge amount of parameters. In 2D-CNN, accuracy is much less sensitive to the batch size, showing
differences around $4\%$, so we may select any batch size or prioritize the choice
based on performance and/or energy criteria.

Table \ref{t:acc-imagenet} shows the accuracy values for ResNetIm and CaffeNet. 
We report top-1 and top-5 ac curacies, where the top-1 is the classic accuracy and the top-5 
is the the percentage of test images for which the correct label is among the five most frequent 
labels considered by the model. On image datasets, the more training data are available, 
the more performance gap in favor of ResNet. Moreover, the large number of parameters allows 
to fit a more discriminant model, overtaking CaffeNet by more than a $10\%$ in Top-5 
and almost $20\%$ in Top-1.

Along batch sizes, all models experience accuracy improvements on larger batches. 
Again, the vast amount of training data available in ImageNet requires larger batch sizes 
to compute more accurate gradients during the training process. On small batch sizes, 
the average gradient of the batch, which is used to update the parameters of the network, 
separates from the mean of the complete training set being more noisy, and therefore, 
worsening updates.

\begin{table}  
\caption{Accuracy on TUM-GAID. We deploy different CNN models and batch sizes in rows, 
and scenarios (temporal and non temporal) in columns. The last column `\textit{AVG}' 
stands for the average of each case weighted by the number of classes. 
Best average results are boldfaced.}\label{t:acc}
\centering
\tabsize
\begin{tabular}{l|c|ccc|ccc|c}
\hline 
Model                   & Batch size &   N  &   B  &   S  &  TN  &  TB  &  TS  &     AVG      \\ 
\hline 
\multirow{3}{*}{ResNet} &         64 & 89.0 & 76.4 & 72.2 & 43.5 & 47.0 & 45.0 & \textbf{76.0}\\ 
                        &        128 & 71.9 & 62.6 & 62.3 & 33.8 & 37.4 & 36.3 &         62.8 \\ 
                        &        256 & 63.5 & 52.7 & 56.1 & 32.8 & 35.5 & 38.1 &         55.4 \\ \hline
\multirow{3}{*}{2D-CNN} &         64 & 95.7 & 87.5 & 87.1 & 45.6 & 47.2 & 47.4 & \textbf{86.0}\\ 
                        &        128 & 95.5 & 84.9 & 83.9 & 51.2 & 41.5 & 51.5 &         84.4 \\ 
                        &        256 & 95.3 & 79.2 & 81.2 & 48.6 & 39.5 & 51.7 &         81.6 \\ 
\hline
\end{tabular} 
\end{table}

\begin{table}  
\caption{Accuracy on ImageNet. CNN models and batch sizes are drawn by rows,
and metrics (top-1 and top-5 accuracy) by columns. Top-1 is the classic accuracy 
and the top-5 is the the percentage of test images for which the correct label 
is among the five labels considered most frequent by the model. Best average 
results are boldfaced.}\label{t:acc-imagenet}
\centering
\tabsize
\begin{tabular}{l|c|ccc|ccc|c}
\hline 
Model                     & Batch size &     Top-1    &        Top-5 \\ \hline
\multirow{3}{*}{CaffeNet} &         64 &         44.0 &         68.7 \\ 
                          &        128 &         52.9 &         76.7 \\ 
                          &        256 & \textbf{57.3}& \textbf{80.4}\\ \hline
\multirow{3}{*}{ResNetIm} &         64 &         61.6 &         84.7 \\ 
                          &        128 &         65.6 &         88.5 \\ 
                          &        256 & \textbf{75.3}& \textbf{92.2}\\ \hline
\end{tabular} 
\end{table}

\subsection{Best approach}\label{s:best-approach}   

We now summarize the information obtained from our experiments to propose 
some guidelines to choose the best hyper-parameters according to performance, 
energy consumption and accuracy criteria.

The road to maximize performance and minimize training time and energy consumption 
leads to larger batch sizes as shown in Tables~\ref{t:edp} and \ref{t:edp-maxwell}. 
On the contrary, when accuracy is a must, small batches should be used on regular datasets. 
Big datasets enable a wider range of batch sizes, and depending on the problem, 
chances to find a good combination of time, energy and accuracy increase. 
When multiple GPUs are available, setups with more than a pair of them must be
carefully studied fas transfers and synchronizations can hurt performance, 
particularly on a set of heterogeneous GPUs.

In general, when small and challenging datasets are used, we have to choose 
between performance/energy or accuracy, but there are exceptions too. For example, 
2D-CNN reaches good accuracy at any batch size. This may indicate that any CNN composed 
solely of 2D convolutions (without batch normalization and residual connections)
can use large batches to save time and energy while retaining accuracy. On the other hand, 
ResNet networks rely on small batches for gaining accuracy at the expense of performance 
and power consumption. Fortunately, this scenario does not predominate and occurs just on 
challenging datasets like TUM-GAID.

On large datasets, state-of-the-art accuracy is obtained with a large batch size 
like 256 samples. In cases like ILSVRC, thanks to the noise effect in the gradients, 
we have it all: optimal accuracy, maximum performance and minimum energy requirements.

In summary, for large datasets the best option is always a large batch size regardless
of the CNN used, and for small datasets, large batches are only useful with networks 
without batch normalization and residual connections. Should a ResNet be used, 
we use a small batch paying a toll in terms of performance and power consumption.
And on a multi-GPU system, it would be convenient to fill all slots available with GPUs alike.


\section{Summary and Conclusions}\label{s:conclusions}

In this paper, we have presented a performance, energy and accuracy analysis 
on a set of popular CNN models running on flagship image and video applications 
for different training sets and parameters setting using the last two 
Nvidia GPU generations, namely Maxwell and Pascal (Titan versions). 
Our goal is to provide an empirical study using state-of-art CNNs 
with applied datasets and carefully selecting parameters of major interest 
for researchers to tune Deep Learning methods. They work on understanding 
how set-ups may help inferences, we evaluate how efficient they are, primarily 
from an energy viewpoint, but also considering speed-ups and numerical accuracy.

Major contributions of this work can be summarized as follows:

\begin{enumerate}

\item We were never able to squeeze more than 55\% of the peak power efficiency 
announced by Nvidia: 20 and 12 GFLOPS/w using the worst-case scenario of 
64-bit data types on Pascal and Maxwell, respectively.

\item The performance per watt gap between Maxwell and Pascal GPUs
was found to reach peaks of up to 60\%, with differences sensitive to the 
CNN model and batch size.

\item If we separate performance and energy, Pascal attains solid differences 
within the $30-40\%$ range depending on the batch size for 2D-CNN, CaffeNet 
and ResNetIm CNN models, in line with Nvidia estimations.

\item Forward and backward steps show similar behaviour in almost all scenarios,
extending performance and power gains on larger CNN batches. 

\item Accuracy prefers small batches on small datasets, but sometimes keeps stable 
on large batches for us to prioritize speed and energy without worsening results.
On big datasets, larger batch sizes minimize trade offs among those three.

\item Datasets play an important role associated to every CNN model, sometimes 
being responsible of inconsistencies and thermal stress in GPU hardware when 
complexity increases. 

\item In multi-GPU environments, the batch size plays an important role because
the GPU code reduces its arithmetical intensity and becomes less compute-bound,
thus requiring larger sizes to hide the communication and synchronization overhead.
In particular, when using four or more GPUs, models trained are required to be huge
for communications to effectively overlap computations. And heterogeneity in 
GPU hardware also introduces additional hurdles for these communication costs.

\end{enumerate}

We envision GPUs to increase their role as high performance and low power devices 
for CNNs and Deep Learning applications in future GPU generations, particularly 
after the introduction of the 3D memory in 2017 and the Volta generation by Nvidia. 
Volta increases the number of cores from 3584 to 5120 to leverage speedups, 
and relaxes frequency from 1480 to 1455 MHz with transistors shrinking from 16 to 12 nm. 
for a more complete low-power device and ambitious GFLOPS/w ratio. Endowing Volta 
with half precision data types and Tensor cores will also affect performance, energy 
and accuracy in a very positive manner, leaving room for a promising scalability 
to constitute the next step of our analysis as future work.

\section*{Acknowledgments}
This work was supported by the Ministry of Education of Spain under Project
TIN2013-42253-P, TIN2016-78799-P (AEI/FEDER, UE) and by the Junta de Andalucia 
under Project of Excellence P12-TIC-1741 and TIC-1692. We thank Nvidia for hardware 
donations within GPU Education Center 2011-2016 and GPU Research Center 2012-2016 
awards at the University of Malaga (Spain). 
We also thank Francisco D. Igual and Luis Pi\~nuel from the Computer Architecture 
and Automated Department at the Complutense University of Madrid (Spain) 
for providing us Accelpower modules to measure power during our experimental survey.
Our measuring system is based on a tool being continuously upgraded as 
reported in http://accelpowercape.dacya.ucm.es.

\bibliographystyle{wileyj}
\bibliography{longstrings,bibliogr}

\end{document}